\documentclass[10pt]{article} % For LaTeX2e

\usepackage[accepted]{rlj} % Should be uncommented for the camera-ready
\usepackage{amssymb}            % Defines common symbols like \mathbb R
\usepackage{mathtools}          % Extends amsmath, providing common math tools
\usepackage{mathrsfs}           % Enables \mathscr, which can work in cases that \mathcal does not
\usepackage{graphicx}           % For including images
\usepackage{subcaption}         % Allows for the use of subfigures and subcaptions
\usepackage[space]{grffile}     % For spaces in image names
\usepackage{url}                % For displaying URLs
\usepackage{lipsum}             % For placeholder text

\usepackage{times}
\usepackage[utf8]{inputenc} % allow utf-8 input
\usepackage[T1]{fontenc}    % use 8-bit T1 fonts
\usepackage{url}            % simple URL typesetting
\usepackage{booktabs}       % professional-quality tables
\usepackage{nicefrac}       % compact symbols for 1/2, etc.
\usepackage{microtype}      % microtypography
\usepackage{adjustbox}      % auto-resizing content 
\usepackage{graphics,color}

\usepackage{algorithm,multicol}
\usepackage[noend]{algpseudocode}
\algrenewcommand\algorithmicindent{1em}
\algrenewcommand{\algorithmiccomment}[1]{%
\bgroup\hskip2em\textcolor{ourdarkgreen}{//~\textsl{#1}}\egroup}

\usepackage{enumitem}

\usepackage{amsfonts}       % blackboard math symbols
\usepackage{amsmath}       % blackboard math symbols
\usepackage{amssymb}
\usepackage{xspace}

\usepackage{xr-hyper}
\usepackage{hyperref}% should be the last package you include except cleveref
\usepackage{cleveref}  % 
\makeatletter
\DeclareRobustCommand\onedot{\futurelet\@let@token\@onedot}
\def\@onedot{\ifx\@let@token.\else.\null\fi\xspace}
\makeatother

\newcommand{\eg}{e.g\onedot}
\newcommand{\ie}{i.e\onedot}

\newcommand{\wrt}{w.r.t\onedot}

\makeatletter
\newcommand*{\addFileDependency}[1]{% argument=file name and extension
  \typeout{(#1)}
  \@addtofilelist{#1}
  \IfFileExists{#1}{}{\typeout{No file #1.}}
}
\makeatother

\usepackage{xcolor}
\definecolor{ourblue}{rgb}{0.368,0.507,0.71}
\definecolor{ourorange}{rgb}{0.881,0.611,0.142}
\definecolor{ourgreen}{rgb}{0.56,0.692,0.195}
\definecolor{ourred}{rgb}{0.923,0.386,0.209}
\definecolor{ourviolet}{rgb}{0.528,0.471,0.701}
\definecolor{ourbrown}{rgb}{0.772,0.432,0.102}
\definecolor{ourlightblue}{rgb}{0.364,0.619,0.782}
\definecolor{ourdarkgreen}{rgb}{0.572,0.586,0.}
\definecolor{thisgreen}{RGB}{97, 163, 58} % the green used in this paper
\definecolor{ourcyan2}{rgb}{0.125,0.722,0.804}
\definecolor{ourred2}{rgb}{0.863,0.184,0.047}
\definecolor{ouryellow2}{cmyk}{0,0.16,1.0,0.07}
\definecolor{ourviolet2}{cmyk}{0.55,0.56,0,0.47}
\definecolor{ourorange2}{cmyk}{0,0.46,0.89,0.11}

\makeatletter
\newcommand{\removeParBefore}{\ifvmode\vspace*{-\baselineskip}\setlength{\parskip}{0ex}\fi}
\newcommand{\removeParAfter}{\@ifnextchar\par\@gobble\relax}
\newcommand{\eq}{\begingroup\removeParBefore\endlinechar=32 \eqinner}
\newcommand{\eqinner}[2][aligned]{\endlinechar=32%
\begin{gather}\begin{#1}#2\end{#1}\end{gather}\endgroup\removeParAfter}
\makeatother

\DeclareDocumentCommand{\p}{ D<>{p} D<>{} r() }{
\def\content{#3}\patchcmd{\content}{|}{\;#2\vert\;}{}{}
\ensuremath{#1 #2(\content #2)}}

\DeclareDocumentCommand{\P}{ D<>{P} D<>{\big} r() }{
\def\content{#3}\patchcmd{\content}{|}{\;#2\vert\;}{}{}
\ensuremath{\operatorname{#1}#2(\content #2)}}

\DeclareDocumentCommand{\E}{ D<>{E} E{_}{{}} D<>{\big} r[] }{
\def\content{#4}\patchcmd{\content}{|}{\;#3\vert\;}{}{}
\ensuremath{\operatorname{#1}_{#2}#3[\content #3]}}

\DeclareDocumentCommand{\D}{ D<>{D} D<>{\big} r[] }{
\def\content{#3}\patchcmd{\content}{||}{\;#2\|\;}{}{}
\ensuremath{\operatorname{#1}\!#2[\content #2]}}

\NewDocumentCommand{\Nor}{ r() }{\P<Normal>](#1)}
\NewDocumentCommand{\Cat}{ r() }{\P<Cat>](#1)}
\NewDocumentCommand{\Bin}{ r() }{\P<Bin>](#1)}
\NewDocumentCommand{\Bet}{ r() }{\P<Beta>](#1)}
\NewDocumentCommand{\Ber}{ r() }{\P<Bernoulli>(#1)}
\NewDocumentCommand{\Dir}{ r() }{\P<Dir>(#1)}

\DeclareDocumentCommand{\KL}{ D<>{\big} r[] }{\D<KL><#1>[#2]}
\DeclareDocumentCommand{\H}{ D<>{\big} r[] }{\E<H><#1>[#2]}
\DeclareDocumentCommand{\I}{ D<>{\big} r[] }{\E<I><#1>[#2]}

\DeclareDocumentCommand{\lnpp}{ D<>{} r() }{
\ensuremath{\p<\ln p_\phi><#1>(#2)}}
\DeclareDocumentCommand{\pp}{ D<>{} r() }{
\ensuremath{\p<p_\phi><#1>(#2)}}
\DeclareDocumentCommand{\qp}{ D<>{} r() }{
\ensuremath{\p<q_\phi><#1>(#2)}}
\DeclareDocumentCommand{\SymLogNormal}{ D<>{} r() }{
\ensuremath{\p<\operatorname{SymLogNormal}><#1>(#2)}}

\newcommand{\sg}{\ensuremath{\operatorname{sg}}}
\usepackage{xcolor}

\definecolor{MK_Two_One}{RGB}{178,24,43} % Nice red
\definecolor{MK_Two_Two}{RGB}{239,138,98}
\definecolor{MK_Two_Three}{RGB}{253,219,199}
\definecolor{MK_Two_Four}{RGB}{209,229,240}
\definecolor{MK_Two_Five}{RGB}{103,169,207}
\definecolor{MK_Two_Six}{RGB}{33,102,172} % Nice blue

\usepackage{hyperref}
\hypersetup{
colorlinks=true
,linkcolor=ourblue
,citecolor=thisgreen
,filecolor=MK_Two_Six
,urlcolor= {blue!80!black}
,menucolor=MK_Two_Five
,runcolor=MK_Two_Four
,linkbordercolor=MK_Two_One
,citebordercolor=MK_Two_Two
,filebordercolor=MK_Two_Three
,urlbordercolor=MK_Two_Six
,menubordercolor=MK_Two_Five
,runbordercolor=MK_Two_Four
}

\usepackage{glossaries}
\usepackage[abbreviations]{glossaries-extra}
\usepackage[utf8]{inputenc}

\glssetcategoryattribute{abbreviation}{indexonlyfirst}{true}

\glssetcategoryattribute{abbreviation}{nohyperfirst}{true}

\newabbreviation{auroc}{AUROC}{Area Under the Receiver Operating Characteristic Curve}
\newabbreviation{accuracy}{Acc}{Accuracy}

\newabbreviation{cnn}{CNN}{Convolutional Neural Network}

\newabbreviation{dmc}{DMC}{Deepmind Control Suite}
\newabbreviation{fov}{FoV}{Field of View}
\newabbreviation{fpr}{FPR}{False Positive Ratio}

\newabbreviation{gnn}{GNN}{Graph Neural Network}
\newabbreviation{gcn}{GCN}{Graph Convolutional Network}
\newabbreviation{gru}{GRU}{Gated Recurrent Unit}
\newabbreviation{gmm}{GMM}{Gaussian Mixture Model}

\newabbreviation{imu}{IMU}{Inertial Measurement Unit}
\newabbreviation{irl}{IRL}{Inverse Reinforcement Learning}
\newabbreviation{iou}{IOU}{Intersection over Union}
\newabbreviation{knn}{KNN}{K-Nearest Neighbors}

\newabbreviation{lagr}{LAGR}{Learning Applied to Ground Vehicles}
\newabbreviation{lidar}{LiDAR}{Light Detection and Ranging}

\newabbreviation{mlp}{MLP}{Multi-Layer Perceptron}
\newabbreviation{mpc}{MPC}{Model Predictive Controller}
\newabbreviation{mse}{MSE}{Mean Squared Error}
\newabbreviation{mae}{MAE}{Mean Absolute Error}
\newabbreviation{mbrl}{MBRL}{Model-based Reinforcement Learning}
\newabbreviation{ood}{OOD}{Out-Of-Distribution}
\newabbreviation{id}{ID}{In-Distribution}
\newabbreviation{rbf}{RBF}{Radial Basis Function}
\newabbreviation{rmp}{RMP}{Riemannian Motion Policies}
\newabbreviation{ros}{ROS}{Robot Operating System}
\newabbreviation{ros1}{ROS~1}{Robot Operating System}
\newabbreviation{roc}{ROC}{Receiver Operating Characteristic}
\newabbreviation{rf}{RF}{Random Forest}
\newabbreviation{rssm}{RSSM}{Recurrent State-Space Model}

\newabbreviation{sdf}{SDF}{Signed Distance Field}
\newabbreviation{slam}{SLAM}{Simultaneous Localization and Mapping}
\newabbreviation{svm}{SVM}{Support Vector Machine}
\newabbreviation{svc}{SVC}{Support Vector Classifier}
\newabbreviation{wvn}{WVN}{Wild Visual Navigation}
\newabbreviation{wm}{WM}{World Model}
\newabbreviation{vit}{ViT}{Vision Transformer}

\newabbreviation{fifo}{FIFO}{First-In-First-Out}
\glsdisablehyper
\graphicspath{{graphics/}}

\usepackage{lineno}
\usepackage{wrapfig}
\usepackage{siunitx}
\usepackage{caption}
\newcommand{\Active}{Active\xspace}
\newcommand{\Passive}{Passive\xspace}
\newcommand{\Tandem}{Tandem\xspace}

\definecolor{ActiveColor}{HTML}{e06363} % Define custom color
\definecolor{PassiveColor}{HTML}{5e81b5} 
\definecolor{TandemColor}{HTML}{5ea339}
\definecolor{y}{HTML}{e19c24}
\definecolor{p}{HTML}{af4bce}

\def\secref#1{Sec.~\ref{#1}}
\def\appref#1{Appendix~\ref{#1}}
\def\figref#1{Fig.~\ref{#1}}

\def\Figref#1{Fig.~\ref{#1}}
\def\tabref#1{Tab.~\ref{#1}}
\def\eqref#1{Eq.~(\ref{#1})}
\def\eqrefp#1{(Eq.~\ref{#1})}
\def\algref#1{Alg.~\ref{#1}}

\title{Offline vs.\ Online Learning in Model-based RL: Lessons for Data Collection Strategies}

\setrunningtitle{Offline vs Online Learning in Model-based RL}

\author{Jiaqi Chen \textsuperscript{1,3}, Ji Shi \textsuperscript{1,2}, Cansu Sancaktar \textsuperscript{1,2}, Jonas Frey \textsuperscript{2,3}, Georg Martius \textsuperscript{1,2}}

\emails{chenjiaq@ethz.ch, \{ji.shi, cansu.sancaktar\}@uni-tuebingen.de, jonfrey@ethz.ch, georg.martius@uni-tuebingen.de}
 
\affiliations{
$^{1}$\textbf{Autonomous Learning Group, Computer Science, University of Tübingen, Germany}\\
$^{2}$\textbf{Max Planck Institute for Intelligent Systems, Tübingen, Germany}\\
$^{3}$\textbf{Robotic Systems Lab, ETH Zurich, Switzerland}\\
}

\contribution{
We provide an in-depth analysis of performance degradation in offline model-based agents with practical considerations. We highlight the coupling of model and policy learning as a primary contributing factor beyond the pure OOD challenge.
    }
    {
In model-free RL, the performance degradation is often linked to limited coverage of offline datasets, which leads to inaccurate value estimates and poor extrapolation of the learned policy~\citep{ostrovski2021difficulty,yue2023understanding,yue2022boosting}. Similar issues plague offline model-based RL~\citep{he2023surveyoffline,MOReL,offlineadapt,mopo,cang2021behavioral}. 
However, an alternative perspective remains overlooked: the influence of data quality and online interaction ratios on the robustness and generalization of world models.
    }

\contribution{
We demonstrate that incorporating exploration data with a mixed reward improves the state-space coverage in offline training. This provides insights in how to create the offline dataset such that the performance degradation can be mitigated and competitive task performance can be maintained.
    }
    {
Existing methods primarily focus on constraining the agent within in-distribution regions for the task~\citep{MOReL,mopo,combo,wang2023coplanner,matsushima2021deploy} but do not explicitly assess which data collection strategies best support offline training.
    }

\contribution{
We propose using the world model loss as a metric to measure the novelty of regions explored by the current policy. It serves as an indicator for when minimal online interactions can help offline agents to efficiently improve performance.
    }
    {
The approach of self-generated data is mostly investigated in the context of model-free RL~\citep{ostrovski2021difficulty,lee2021offlinetoonline}.
    }

 % Your keywords

\summary{Data collection is crucial for learning robust world models in model-based reinforcement learning.
The most prevalent strategies are to actively collect trajectories by interacting with the environment during online training or training on offline datasets.
At first glance, the nature of learning task-agnostic environment dynamics makes world models a good candidate for effective offline training. However, the effects of online vs. offline data on world models and thus on the resulting task performance have not been thoroughly studied in the literature. In this work, we investigate both paradigms in model-based settings, conducting experiments on 31 different environments.
First, we showcase that online agents outperform their offline counterparts.
We identify a key challenge behind performance degradation of offline agents: encountering Out-Of-Distribution (OOD) states at test time.
This issue arises because, without the self-correction mechanism in online agents, offline datasets with limited state space coverage induce a mismatch between the agent's imagination and real rollouts, compromising policy training.
We demonstrate that this issue can be mitigated by allowing for additional online interactions in a fixed or adaptive schedule, restoring the performance of online training with limited interaction data.
We also showcase that incorporating exploration data helps mitigate the performance degradation of offline agents. Based on our insights, we recommend adding exploration data when collecting large datasets, as current efforts predominantly focus on expert data alone.
}

\begin{document}

% \makeCover  % Create the cover page
\maketitle  % Make the title section

\begin{abstract}
Data collection is crucial for learning robust world models in model-based reinforcement learning.
The most prevalent strategies are to actively collect trajectories by interacting with the environment during online training or training on offline datasets.
At first glance, the nature of learning task-agnostic environment dynamics makes world models a good candidate for effective offline training. However, the effects of online vs. offline data on world models and thus on the resulting task performance have not been thoroughly studied in the literature. In this work, we investigate both paradigms in model-based settings, conducting experiments on 31 different environments.
First, we showcase that online agents outperform their offline counterparts.
We identify a key challenge behind performance degradation of offline agents: encountering Out-Of-Distribution states at test time.
This issue arises because, without the self-correction mechanism in online agents, offline datasets with limited state space coverage induce a mismatch between the agent's imagination and real rollouts, compromising policy training.
We demonstrate that this issue can be mitigated by allowing for additional online interactions in a fixed or adaptive schedule, restoring the performance of online training with limited interaction data.
We also showcase that incorporating exploration data helps mitigate the performance degradation of offline agents. Based on our insights, we recommend adding exploration data when collecting large datasets, as current efforts predominantly focus on expert data alone.
\end{abstract}
\fancyhead[R]{}
\section{Introduction}\label{sec:intro}
\vspace{-0.5em}

Online training of reinforcement learning (RL) agents enables continual adaptation through direct interaction with the environment. However, this approach is often impractical and less scalable in real-world settings due to high data collection costs, safety concerns, or hardware constraints~\citep{geneoffline,costly,safedreamer}. To address these limitations, offline RL methods attempt to reuse past experiences, training agents on pre-collected datasets without further environment interaction.

However, offline RL is prone to performance degradation when encountering \gls{ood} states, where poor generalization manifests as overestimation errors in value functions, leading to suboptimal action choices~\citep{ostrovski2021difficulty,yue2023understanding,yue2022boosting}. \gls{mbrl} offers a potential alternative by learning task-agnostic environment dynamics, enabling agents to train policies via model rollouts instead of direct environment interaction \citep{bruce2024genie}. 
In principle, this should help mitigate overestimation errors in value functions and thus promote generalization.

Yet, recent studies have shown that MBRL is still vulnerable to \gls{ood} issues, particularly when world models are trained on offline data with insufficient state-space coverage~\citep{mopo, MOReL, wang2023coplanner}, which in turn increases the risk of inaccuracies in the world model—a second source of error distinct from that in value function estimation. 
While prior works, such as MOPO~\citep{mopo}, focus on mitigating distributional shift by penalizing model uncertainty during policy deployment, our study takes a fundamentally different approach: rather than addressing \gls{ood} errors post hoc, we investigate how data diversity, dataset optimality, and online interaction ratios impact the robustness (\ie generalization ability) of world models and \gls{mbrl} policies.
By systematically decoupling the roles of the policy and the world model, we aim to provide a deeper understanding of the failure modes of \gls{mbrl} in offline settings. Our study shifts the focus from uncertainty-penalization techniques to data-driven solutions, offering insights into how data collection strategies influence the reliability and generalization of world models.

In this work, we aim to provide an exhaustive analysis of online and offline data collection paradigms in an \gls{mbrl} setting and address two key questions:
(1) How can we best \textbf{leverage offline data} to train a robust world model and (2)
\textbf{what combination of data collection strategies}—\eg, online and offline, task-oriented and exploration-driven—yields the best performance at the lowest cost across different scenarios? We believe this is a crucial research direction, as analyzing these phenomena from a unified perspective across a wide range of environments can provide valuable insights for future dataset collection.

\begin{figure}[t]
   \centering
   \includegraphics[scale=1.0]{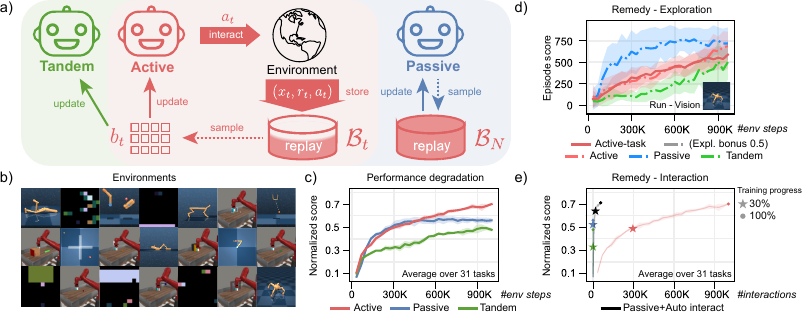}
   \vspace{-.2em}
   \caption{\textbf{Investigation of the performance degradation in offline agents and potential remedies.} a)~Illustration of \Active, \Passive, and \Tandem agents. The \Active agent is trained using online RL and is allowed to interact with the environment.
   The \Passive agent is trained from the full buffer of an \Active agent, without performing any additional interactions. The \Tandem agent, is also trained offline, but samples batches from the \Active agent's replay buffer in the exact same sequence.
   b)~We conduct experiments in 31 tasks across various domains.
   c)~Illustration of the performance degradation in \Passive and \Tandem agents \wrt the \Active agent.
   d-e)~exploration data~(d) and online interaction~(e) effectively mitigate performance degradation observed in offline \Passive agents.\looseness-1}
   \label{agents}
   \vspace{-0.7em}
\end{figure}

We employ DreamerV3~\citep{hafner2023mastering} across 31 diverse environments on well-established benchmarks including locomotion, manipulation, and numerous other robotic tasks. As shown in \figref{agents}, we examine three scenarios: (1)~an \Active agent training tabula rasa, (2)~a \Tandem agent replaying the learning history of the \Active agent in the same temporal order but with a different random initialization, and (3)~a \Passive agent with access to the \Active agent's full experience from the start, also with a different random initialization.
Importantly, both the \Tandem and \Passive agents do not collect data themselves; instead, they learn offline from data generated by the \Active agent.

Our key findings reveal that in a task-oriented setting\footnote{That is, an agent trained solely with task-specific rewards.}, \Tandem and \Passive agents underperform compared to the \Active agent, primarily due to visiting novel states during evaluation.
This \gls{ood} tendency stems from the absence of a self-correction mechanism in offline agents.
Unlike online agents, which can correct overestimation bias through direct interaction with the environment, offline agents cannot gather relevant data to correct their predictions. 
This leads to a mismatch between the agent’s imagination and real rollouts, ultimately misguiding policy training.
We demonstrate that using offline exploration data instead of solely task-oriented data mitigates this problem and, surprisingly, find that expert demonstrations alone are insufficient for high performance in \gls{mbrl}.
However, we showcase that performance can be recovered with minimal environment interactions. 
Based on these results, we analyze an adaptive fine-tuning agent that can recover the \Active agent's performance with just \SI{6}{\%} of environment interactions relative to its offline dataset.
As a result of our large-scale experimental study, we suggest to everyone collecting expert demonstration data to also collect exploration data for sufficient state-space coverage.

Our contributions are as follows:
\begin{itemize}[topsep=0em]
    \item \textbf{Analysing the process behind performance degradation} in offline model-based agents, along with several practical considerations.
    \item \textbf{Demonstrating the benefits of exploration data} and proposing that a mixed reward function enhances state-space coverage in data collection, preventing performance degradation in offline training while maintaining strong task performance.
    \item \textbf{Examining world-model loss as a metric for targeted active data collection}, thereby substantially enhancing the efficiency of offline agents with minimal additional interactions.
\end{itemize}

\section{Method}
\subsection{Preliminaries}\label{sec:preli}
\paragraph{Model-based Reinforcement Learning}\label{sec:mbrl}

In this work, we consider environments that can be described by a partially observable Markov Decision Process (POMDP), with high-dimensional observations $x_t$, which are encoded into latent representations $s_t$, state-conditioned actions $a_t$ generated by an agent and scalar rewards $r_t$ (conditional on $s_t$ and  $a_t$) generated by the environment.
In \gls{mbrl}, our aim is to learn the latent transition dynamics by a \textbf{world model} $\hat{\mathcal{T}}(s_{t+1} \mid s_t, a_t)$ and find an optimal \textbf{policy} $\pi(a_t|s_t)$ maximizing the expected discounted return with discount factor $\gamma$:
\begin{equation}
\label{mbrl-goal}
\pi^* = \mathop{\arg\max}_{\pi} \mathop{\mathbb{E}}_{\substack{s_{t} \sim \hat{\mathcal{T}}(\cdot \mid s_{t-1}, a_{t-1}) \\ a_t \sim \pi(a \mid s_t)}} \left[\sum_{t=0}^{\infty} \gamma^t r(s_t, a_t)\right].
\end{equation}

\paragraph{DreamerV3}

We use DreamerV3~\citep{hafner2023mastering}, a state-of-the-art model-based RL method, as the base architecture in all our experiments.
Based on the \gls{rssm}~\citep{hafner2018planet} summarized in \eqref{eq:wm}, the world model predicts the latent state $s_t = (h_t, z_t)$ from the previous state and action, where $h_t$ is the deterministic and $z_t$ is the stochastic state component.
The estimated observation $\hat{x}_t$, reward $\hat{r}_t$, and continuation flag $\hat{c}_t $ (signalling whether the episode has ended or not) are decoded from the latent states; given by the tuple $\hat{e}_t = ( \hat{x}_t, \hat{r}_t, \hat{c}_t)$.
The policy has an actor-critic architecture, detailed in \eqref{eq:ac}.
$R_t$ is the discounted return from state $s_t$.
For the off-policy updates of DreamerV3, environment interactions are added to a replay buffer $\mathcal{B} = \{ (x_t, a_t, r_t, c_t, \dots)\}_{t=1}^{N}$, where each tuple contains the observation $x_t$, action $a_t$, reward $r_t$,  continuation flag $c_t$, and optionally other variables collected from the environment.

\eq{
\begin{alignedat}{2} \textbf{Sequence model:} \quad & h_t = f_\phi(h_{t-1}, z_{t-1}, a_{t-1}) \quad\quad & \textbf{Encoder:} \quad & z_t \sim \qp(z_t \mid h_t, x_t) \quad\quad \\
\textbf{Dynamics predictor:} \quad & \hat{z}_t \sim \pp(\hat{z}_t \mid h_t)
 & \textbf{Decoder:} \quad & \hat{e}_t \sim \pp(\hat{e}_t \mid h_t, z_t)
\end{alignedat}
\label{eq:wm}
 }

\eq{
&\textbf{Actor:}\quad
&& a_t \sim \p<\pi_\theta>(a_t|s_t) \quad\quad\quad\quad
&\textbf{Critic:}\quad
&& v_\psi(s_t) \approx \E_{p_\phi,\pi_\theta}[R_t]
\label{eq:ac}
}

DreamerV3 minimizes the world model loss, which is a weighted loss of multiple components and is defined in the original paper~\citep{hafner2023mastering}, as shown in \eqref{eq:wm_loss}.

\eq{
\mathcal{L}(\phi)\doteq
\E_{q_\phi}<\Big>[\textstyle\sum_{t=1}^T(
    \beta_{\mathrm{dyn}}\mathcal{L}_{\mathrm{dyn}}(\phi)
   +\beta_{\mathrm{rep}}\mathcal{L}_{\mathrm{rep}}(\phi)
   +\beta_{\mathrm{pred}}\mathcal{L}_{\mathrm{pred}}(\phi)
)].
\label{eq:wm_loss}
}

It consists of the dynamics-based loss components given by $\mathcal{L}_{\mathrm{dyn}}$ and $\mathcal{L}_{\mathrm{rep}}$, defined in \eqref{eq:wm_loss-comps}, as well as the loss $\mathcal{L}_{\mathrm{pred}}$ from three prediction heads: observation reconstruction, reward estimation, and continuity prediction.

The following three-step cycle is repeated throughout the training process of DreamerV3:
(1)~The agent interacts with the environment to collect data, adding it to its replay buffer $\mathcal{B}$.
Meanwhile, the latent states $(h_t,z_t)$ are updated closed-loop using the current observation $x_t$ and are used to compute the action.
(2)~The world model is trained on a batch of sequence data uniformly sampled from the replay buffer using the loss function shown in \eqref{eq:wm_loss}.
(3)~Open-loop trajectories are generated in imagination by the world model to train the actor and critic networks.

\subsection{Learning Agents}\label{sec:regimes}

In order to investigate the online and offline training paradigms, we design three off-policy agents, as shown in \figref{agents}, each representing a different variation of training data collection.

\textbf{\Active agent} is the typical RL agent in online RL.
It interacts with the environment and performs training steps using the collected data by its own policy.
An \Active agent can adapt its world model with its own policy rollouts, which is a self-correction mechanism, enabling the agent to learn from its own mistakes~\citep{ostrovski2021difficulty}.

\textbf{\Passive agent} is trained offline without any environment interactions by uniformly sampling data from the \emph{final} replay buffer $\mathcal{B}_N$ of an \Active agent.
This gives the \Passive agent access to the full data of the \Active agent right from the start of the training process, including high-reward trajectories.
It serves as a conventional offline agent, trained on static data without interaction or replay dynamics.

\textbf{\Tandem agent} is another agent trained offline, but sees the training data in the same order as the \Active agent, \ie the training batches $b_t$ are replayed exactly as they were sampled during the training of the \Active agent~\citep{ostrovski2021difficulty}. 
The goal here is to introduce a more controlled offline learning setting than the \Passive agent, with the only difference from the \Active agent being the model initialization.
This setup facilitates easier interpretation of the experimental results.

The offline agents, \Passive and \Tandem, are initialized independently of the \Active agent used for data collection with a different random seed. The pseudocode of the agents is in \appref{subsec:pseudocode}.

\vspace{-0.5em}
\section{Experiments}\label{sec:exp-analysis}
\vspace{-0.5em}
We use DreamerV3 for all our experiments (details on hyperparameters can be found in \appref{sec:hyperparam}).
In total, we conducted \num{2000} experiments using \num{20000} GPU hours. All agents are trained from scratch using task-oriented rewards unless specified otherwise.

\subsection{Environment Setup}\label{sec:setup}

Our experiments are conducted in the \gls{dmc}~\citep{dmc,yarats2022exorl}, Metaworld~\citep{yu2019meta}, and MinAtar~\citep{young19minatar} domains, including a total of 31 tasks.
These are representative environments for robotic locomotion, manipulation, and discrete game tasks.
The environment settings mainly follow the default settings in~\citet{hafner2023mastering}.
The results for all individual experiments and detailed setups are provided in the \appref{app:complete} and \appref{sec:hyperparam}.
Whether state or image observations are used is indicated alongside the task name as ``proprio'' or ``vision'' respectively.
We run 1 million environment steps per task, training every second step.
Results are averaged over three random seeds and reported as the mean with a shaded region indicating $\pm 1$ standard deviation, unless stated otherwise.
For the \Passive and \Tandem agents, we keep the same total number of environment and training steps as the \Active agent to ensure consistency and comparability; however, without collecting any interaction data, as explained in \appref{subsec:explain-envsteps}.

\subsection{Metrics for Analysis}\label{sec:met}

\textbf{World model loss\quad}
The mean error of the world model for the prediction of dynamics, observation, reward, and continuity (\secref{sec:mbrl}).
It is an indicator of the total aleatoric and epistemic model uncertainty and can serve as a simple \gls{ood} measure~\citep{mopo,offlineadapt}.

\textbf{Episode score\quad}
The undiscounted sum of rewards over the episode.

The metrics shown in all figures are calculated as follows, unless specified otherwise: (1)~Every 5K environment steps, we roll out the agent's policy for a total of 4 episodes. (2)~We compute the mean episode score and the mean world model loss across the 4 episodes (see \secref{subsec:eval-details} for implementation details).
Each agent is evaluated in an on-policy manner on its own test-time trajectories.
The data distributions of visited states are thus conditioned on the policy and are different for individual agents.

\subsection{Toy Example}\label{sec:toy}
\vspace{-.2em}
\begin{figure}[tb]
   \centering
   \includegraphics[scale=1.0]{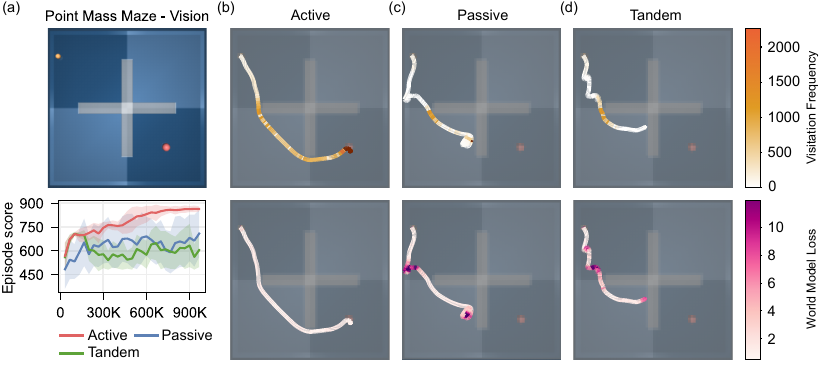}
   \vspace{-1.8em}
   \caption{\textbf{Example of the degraded performance during offline training in 2D point mass maze environment.} The task is to move the yellow point mass from the top-left initial position to the red marker in the bottom-right of the maze, which is the goal position. The episode score of each agent is shown in (a). In (b-d), we show the point mass trajectory generated by the final model after 1M environment steps. The two heatmaps on the trajectory represent: (1)~a count-based frequency of each covered cell that is visited in the replay buffer and (2)~world model loss on each visited state. The median visitation frequency along the shown trajectory is 650.0 for \Active, 54.5 for \Passive, and 37.0 for \Tandem.}
   \label{toy}
   \vspace{-1.5em}
\end{figure}
We first study the performance of all learning agents in a toy environment.
We select the point mass maze environment in \gls{dmc}, where an actuated 2-DoF point mass has to reach the red goal position, as shown in \figref{toy}.
The results show that only the \Active agent successfully solves the task, while both agents trained offline fail, showing degraded performance compared to the \Active agent.

\textbf{Hypothesis: Lack of self-correction causes \gls{ood} errors\quad}The policy in DreamerV3 is trained purely in the imagination of the world model.
As a result, the policy can learn to exploit inaccuracies in the imagination.
The \Active agent continuously collects data from regions where the world model could be unreliable, specifically for regions where the world model predicts a high reward and, therefore, the policy is likely to visit. 
Training the world model on the collected data from these regions helps to improve the world model in a targeted manner with respect to the current \Active agent's policy. 
This not only helps to improve the policy to solve the task but also makes the world model adapt to the agent's policy rollouts, ensuring sufficient data coverage around its self-rollouts. Consequently, the agent is unlikely to encounter novel states when rolling out the policy during evaluation. 

The agents trained offline lack this critical feedback loop of self-correction.
Although the overall training data distribution is the same as the \Active agent, differences in sampling sequences (\Passive) and/or model initializations (\Passive and \Tandem) lead to distinct policies during training. 
To effectively improve these policies, the training data generated from the world model's imagination should closely match real rollout performance. 
However, without self-correction and constrained by data coverage tailored to another agent’s policy, the imagination of this limited-capability world model fails to align with real rollouts under its own policy, leading to a persistent discrepancy between imagination and reality in offline training. 
Consequently, the policy will exploit these inaccuracies during training and be updated blindly to eventually steer the agent toward novel, unvisited areas.
During test time, visiting novel states can lead to world model prediction errors and, therefore, suboptimal policy actions.
It creates a catastrophic cycle where each compromised action leads to further novel states and additional inaccuracies in the world model until the episode ends or the agent accidentally re-enters into a familiar state.

We observe this behavior in the performance of the three agents as shown in \figref{toy}. 
The \Active agent learned to adapt its world model to its own rollouts; therefore, it did not meet any novel states when rolling out the policy for evaluation, as shown by the consistent low world model loss and high visitation frequencies alongside its trajectory. 
However, this is not the case for the \Passive and \Tandem agents. From the start, their policies seem to behave anomalously, guiding them towards a suboptimal direction even in the regions familiar to the world model. 
This is due to the absence of feedback: the policy exploits model inaccuracies, and blind updates cause it to drift further from optimal behavior over time.
Since the task-oriented dataset has limited state-space coverage, they inevitably visit novel states.
Although both the \Passive and \Tandem agents re-enter familiar states under arbitrary actions, their compromised policies subsequently lead them to another \gls{ood} region, where they cannot recover until the end of the episode—ultimately failing to solve the task.\looseness=-1

To summarize, \textbf{self-correction ensures sufficient data coverage related to the agent's policy rollouts}, thereby \textbf{(1)~preventing \gls{ood} errors} and \textbf{(2)~facilitating policy training} by reducing gaps between imaginations and real rollouts. Without self-correction, imagination gaps compromise policy training and push offline agents toward \gls{ood} states, where they become trapped in a catastrophic cycle that leads to further performance degradation.\looseness=-1

Our hypothesis is generally in line with previous research in model-free RL~\citep{ostrovski2021difficulty,yue2023understanding,emedom2023knowledge,kumar2020conservative}
, which attributes performance degradation to extrapolation errors in Q-values in \gls{ood} state-action pairs during training and evaluation.
However, in the context of \gls{mbrl}, the paradigm is shifted from a focus on Q-functions to the coupling of a world model and a policy network.\looseness=-1

\vspace{-.6em}
\subsection{Validation across Tasks}\label{sec:val-tasks}
\vspace{-.3em}
\begin{figure}[tb]
   \centering
   \includegraphics[scale=.95]{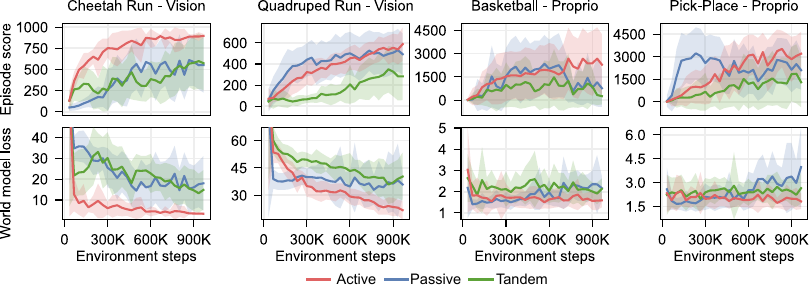}
   \vspace{-1.1em}
   \caption{\textbf{Episode score and world model loss during evaluation rollouts of 4 selected tasks.} The first two are from \gls{dmc} and the last two are from the Metaworld domain. The performance degradation of offline agents, including \Passive and \Tandem, is common across domains and tasks, especially for \Tandem agents.}
   \label{episode_score_wmloss_4tasks}
   \vspace{-1.8em}
\end{figure}

The performance degradation phenomenon in offline agents is observed across various tasks and domains, as shown in \figref{episode_score_wmloss_4tasks} and \appref{app:complete-task}.
In tasks such as \emph{Quadruped Run - Vision} and \emph{Pick-Place - Proprio}, the \Passive agent initially demonstrates a faster increase in performance but has a larger variance or even experiences performance drops as training progresses.
The degraded performance in \Passive and \Tandem agents is accompanied by a significantly larger world model loss on evaluation episodes than the \Active agent.
Given that a high world model loss indicates novel states, this observation supports our hypothesis in \secref{sec:toy}.
The discrepancy between imagined and real rollouts in offline agents is shown in \appref{sec:discrepancy}.
Our detailed inspections on a timestep level in \appref{sec:step-analysis} further validate our hypothesis of the catastrophic cycle during testing.
\figref{episode_score_wmloss_4tasks} also shows a potential advantage of \Passive agents: faster convergence by having access to high reward trajectories from the start of training (validated in \appref{sec:ood-app-detail}), though additional measures may be necessary to ensure training stability.
The results of \Tandem agents also follow the findings of degraded performance of the \Tandem training regime in~\citet{ostrovski2021difficulty} and extend its validity to \gls{mbrl}.
We include more discussions about the \Tandem agent and the self-correction mechanism in~\appref{sec:more-self-correction}.\looseness-1

\subsection{Deep Dive into Performance Degradation}\label{subsec:deep-dive}

\subsubsection[OOD in MBRL]{\gls{ood} in \gls{mbrl}}\label{subsubsec:why-ood}

\begin{figure}[t]
   \centering
    \includegraphics[scale=0.9]{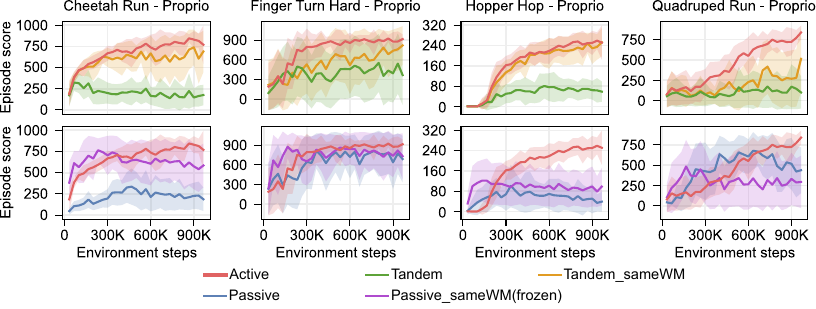}
   \vspace{-.7em}
   \caption{\textbf{Performance comparison when keeping an equivalent world model in \Passive or \Tandem agents to the one of the \Active agent} throughout training. Despite utilizing the same world model during training, performance degradation still occurs, albeit to varying degrees.}
   \label{wmsame}
   \vspace{-8pt}
\end{figure}

\textbf{Both world model and policy affect performance degradation}
To decouple the effect of the world model and the policy on the performance degradation, we carry out a more controlled experiment as shown in \figref{wmsame}.
In this setup, the \Tandem agent's world model replicates that of the \Active agent precisely at each training step, which is referred to as Tandem\_sameWM. For \Passive agents, we keep using the final world model from their \Active counterpart for the remainder of training, which is named Passive\_sameWM(frozen). The pseudocode of the agents is in \appref{subsec:pseudocode}. \looseness=-1

After isolating the effect of different world models, we observe that the degradation still persists although the extent of it varies across tasks. 
In tasks such as \emph{Hopper Hop - Proprio}, the performance degradation of the Tandem\_sameWM agent is minimal, while it remains significant in others like \emph{Quadruped - Proprio}. 
A similar trend appears in Passive\_sameWM(frozen) agents. 
These findings suggest that deviations in both the world model and policy from the \Active agent contribute to performance degradation, with their relative impacts depending on the specific task.
More discussions about this experiment can be found in~\appref{sec:more-self-correction}.\looseness=-1

\textbf{What is the difference to supervised learning?\quad}
In classical supervised learning, a model is optimized on an offline dataset, \eg, for image classification. Training on independent and identically distributed data from different random initializations typically yields similar performance, showing robustness to initialization.
Why is this not the case in the \gls{mbrl} setting, where \Tandem agents perform worse than \Active agents, despite one expecting the world model to perform equally well across seeds given the same data?
This is because offline trained agents will cause states to be visited during policy optimization that are not collected by the Active agent, leading to OOD queries to the model.\looseness-1

\subsubsection{World model loss is a pessimistic indicator of performance degradation}\label{subsubsec:pessim}
The world model loss is due to prediction errors arising from both epistemic and aleatoric uncertainty. 
Novel states lead to high variance predictions due to epistemic uncertainty induced by insufficient state space coverage during training.
Overlaid are errors due to partial observability and environment stochasticity.

\begin{wrapfigure}[16]{r}{.38\textwidth}
    
   \centering
   \includegraphics[scale=.9]{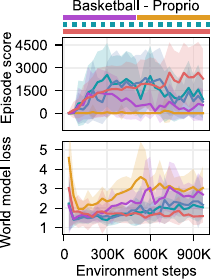}
   \caption{\textbf{Performance comparison of \textcolor[HTML]{e06363}{\Active}, \textcolor[HTML]{5e81b5}{\Passive} as well as Passive agents trained on \textcolor[HTML]{e19c24}{expert}, \textcolor[HTML]{af4bce}{suboptimal}, and  \textcolor[HTML]{1298a5}{mixed} data}, which is implemented by splitting the replay buffer of the \Active agent in different ways.
   }
   \label{expert_small}
\end{wrapfigure}

In particular, the latter factors can lead to high model loss without significant impacts on performance,
depending on whether exact predictions are required for the task at hand. 

In addition, even when the agent is in novel states, other factors, \eg environment constraints, and the policy producing correct actions by coincidence in hallucinations of the world model, can reduce the impact of a poorly performing world model on agent performance.
Therefore, the world model loss is a pessimistic indicator of performance degradation.

\subsubsection{Expert data alone exacerbates OOD issues}\label{subsubsec:overfit}

Expert data is commonly used in offline learning, but compared to data collected by the \Active agent, its coverage is more limited to task-specific trajectories, typically capturing only certain ways of solving the task. As a result, states are more likely to be \gls{ood} for the world model, resulting in even worse task performance, as shown in \figref{expert_small}, where we treat the second half of the buffer as expert data.
As expected, the world model loss evaluated on test-time trajectories is significantly larger than for other agents.
For more details, see \appref{sec:ood-app-detail}.

\subsubsection{Considerations in Practical Applications}
\label{sec:ood-app}

In further experiments, we find that initializing the \Passive agents' weights identically to the \Active agents' does not improve task performance.
In contrast, even minor differences in the model initialization of \Tandem agents compared to \Active agents leads to degraded performance, reflecting the chaotic training dynamics of gradient-based optimization.
See \appref{sec:ood-app-detail} for more details.

\vspace{-0.7em}
\section{Potential Remedies from a Data Perspective}\label{sec:remedy}

Based on the previous analysis, we conclude that insufficient state coverage during training of \Passive and \Tandem agents limits the generalization capability of the world model, which results in the policy exploiting inaccuracies of the world model during training and eventually visiting \gls{ood} states during evaluation.
To address this, we propose two strategies for effective agent training with offline datasets: \textbf{training on an exploration dataset} and \textbf{(adaptively) incorporating self-generated data}.\looseness-1

\subsection{Training on Exploration Data}\label{sec:exploration}

\begin{figure}[b]
   \centering
   \vspace{-1.5em}
   \includegraphics[scale=1.0]{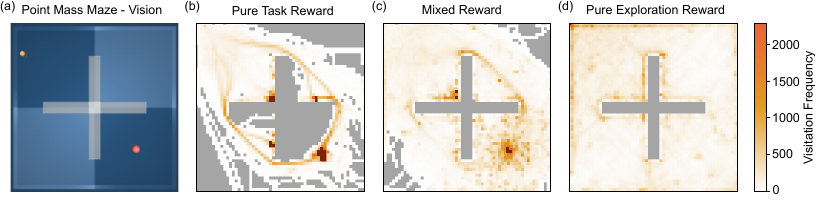}
   \vspace{-.7em}
   \caption{\textbf{State visitation in the Point Mass Maze task.} They are calculated using the discretized states from three different \Active agents' final replay buffers after 1M environment steps. 
   (b)~Agent in a pure task-oriented setting. 
   (c)~Agent with a mixed reward: task plus exploration rewards, see \eqref{eq:reward_bonus} with $w_\mathrm{expl}=0.5$. 
   (d)~Agent with pure exploration rewards based on ensemble disagreement~\citep{sekar2020planning}. 
   The unvisited areas are painted gray, and the outliers that have extremely high values are painted dark red. 
   Here the task-oriented agent only explores limited state space in the map and always follows certain routes towards the goal position, while the two explorative agents visit the space much more equally.}
   \label{heatmap}
   \vspace{-5pt}
\end{figure}
\begin{figure}[thb]
   \centering
   \includegraphics[scale=1.0]{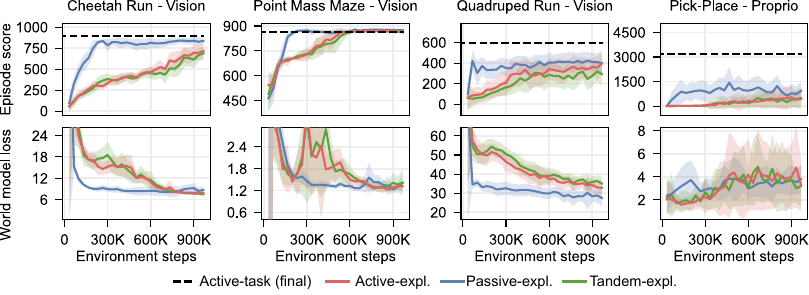}
   \vspace{-.7em} % there is too much space here, for the caption should be clear where it belongs to.
   \caption{\textbf{Performance comparison when training on pure exploration data.} The dataset is generated by the Active-expl. agent with a behavioral policy based on ensemble disagreement~\citep{sekar2020planning}. We additionally show the baseline performance of a task-oriented \Active agent.}
   \label{explore_100}
   \vspace{-5pt}
\end{figure}
We investigate how training on exploration data affects the performance of \Active, \Passive and \Tandem agents. 
Here, we use Plan2Explore~\citep{sekar2020planning}, where the objective is to maximize the information gain of the world model. The exploration reward is calculated as ensemble disagreement, denoted by $r_\mathrm{disag}$.
We investigate exploration in two modes: (1)~pure exploration in a task-free setting, \ie the agent only maximizes for $r_\mathrm{disag}$, (2)~a mixed reward setting, where $r_\mathrm{disag}$ is added as an exploration bonus on top of the task reward: 
\eq{ 
r_t & \doteq
w_\mathrm{task} \cdot r_\mathrm{task}+ w_\mathrm{expl} \cdot r_\mathrm{disag},
\label{eq:reward_bonus}
}
where $w_\mathrm{task}$ and $w_\mathrm{expl}$ weights are normalized such that they sum up to 1.
In both cases, the \Active agent trains two separate policies: an exploration policy (guided by either pure exploration or the mixed reward, used to collect data), and a task policy (trained only for task evaluation).

For agents trained offline, exploration data in the training set can provide a larger state-space coverage, which can counteract the missing self-correction mechanisms of an active agent.
\Figref{heatmap} demonstrates how task-oriented data is narrower compared to exploration data.
The addition of exploration data becomes crucial in alleviating the \gls{ood} challenge during evaluation, as validated in \figref{explore_100}, where the training data is gathered by an \Active agent based on pure exploration rewards $r_\mathrm{disag}$.
As a result, the \Passive agents generally outperform their \Active counterparts, and in some tasks even match the performance of the online task-oriented version, while the \Tandem agents perform comparably to the \Active agents in most cases.
Furthermore, the relationship between task performance and world model loss generally also matches the findings in \secref{sec:val-tasks}.
However, some cases in \appref{app:complete-expl} indicate that world model loss can occasionally be less predictive of task performance.
This inconsistency arises as novel regions for the world model shrink with exploration data, leading to lower world model loss—even in regions far from typical task trajectories (more discussions in \appref{app:outlier-complete-res}).
In addition, the pure exploration dataset contains numerous trajectories irrelevant to the task, interfering with the effective learning of the task policy.
Consequently, task performance becomes increasingly dependent on the task difficulty.
For example, in two challenging tasks~--~\emph{Quadruped Run - Vision} and \emph{Pick-Place - Proprio}~--~
agents trained on pure exploration data have significantly lower performance than those trained with task-oriented data, as shown in \figref{explore_100}.\looseness-1

\begin{figure}[tb]
   \centering

   \includegraphics[scale=1.0]{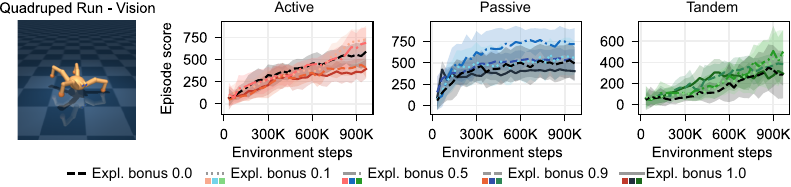}
   \vspace{-.7em}
   \caption{\textbf{Training on pure exploration data is not optimal.} Performance comparison when assigning different exploration bonuses $w_\mathrm{expl}$ in the reward function. The black dashed lines represent pure task-oriented policy without any exploration bonus.}
   \label{explore_bonus}
   \vspace{-1.5em}
\end{figure}

To this end, we investigate the mixed reward setting, where we add the exploration reward as a bonus, as defined in \eqref{eq:reward_bonus}.
This approach allows a more concentrated exploration near the goal, as shown in \figref{heatmap}, preventing the excessive exploration of irrelevant areas that could arise from a purely explorative dataset.

Indeed, in \figref{explore_bonus}, we show that pure exploration is hardly the best option for the hard tasks like \emph{Quadruped Run - Vision}.
The addition of an exploration bonus \eg $w_\mathrm{expl}=0.5$ together with task rewards in \emph{Quadruped Run - Vision} can lead to an improved task performance compared to runs with pure task rewards, especially in \Passive agents.
A downside of this approach is the introduction of the hyperparameter $w_\mathrm{expl}$, the optimal value of which can depend on the specific task as shown in our experiments in \appref{app:complete-expl-bonus}.
\vspace{-1em}
\subsection{Adding Additional Self-generated Data}\label{sec:self-interact}
\vspace{-0.5em}
\begin{figure}[tb]
   \centering
   \vspace{-1.5em}
   \includegraphics[scale=1.0]{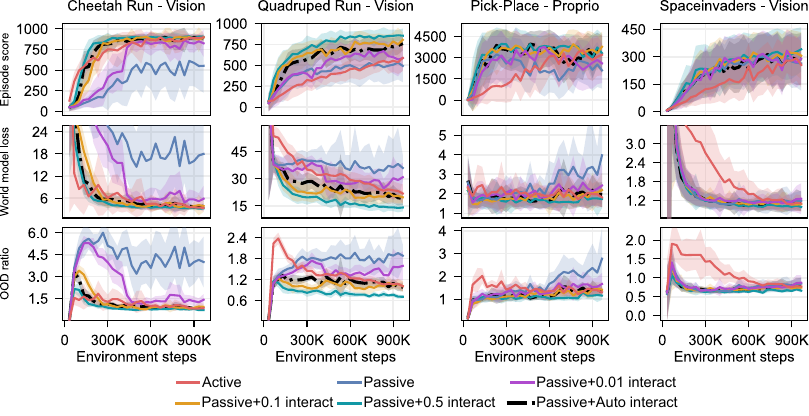}
   \vspace{-.7em}
   \caption{\textbf{Performance comparison when allowing adding additional self-generated data for \Passive agents.} The Passive+Auto interact agent adds 6.5\% self-generated data in Cheetah Run - Vision, 2.9\% in Quadruped Run - Vision, 9.8\% in Pick-Place - Proprio, and 0.5\% in Spaceinvaders. The percentage is calculated \wrt to the size of the final replay buffer of \Active agents.}
   \label{interact}
   \vspace{-1.5em}
\end{figure}
We have demonstrated the critical importance of self-correction. However, as training solely on interaction data is expensive, and offline data is often cheaply available; we would like to explore how one can most effectively combine fixed offline data with online interaction data. To analyze this interplay, we first examine a strategy that uses a predetermined schedule for the \Passive agent to interact with its environment.

Specifically, for every $N$ environment steps, the \Passive agent is allowed to collect 2K-step transitions based on its learned policy. Then the interactive data will be added to expand the replay buffer for later sampling during world model training as usual.
By choosing a different $N$, we can adjust the frequency of interactive data injection.
Experiments were conducted with $N$ set to 4K, 20K, and 200K, respectively corresponding to 50\%, 10\%, and 1\% self-generated data.
The results are shown in \figref{interact}. Accordingly, merely 10\% additional self-generated data can already result in a significant improvement in the episode score as well as a notable reduction in the world model loss, recovering the performance of its \Active counterpart.
In certain environments, such as the \emph{Spaceinvaders} from the MinAtar domain, the \Passive agents may already solve the task and have a faster convergence than the \Active one; therefore, self-generated data provides no performance increase.

\textbf{Adaptive interaction\quad} 
Upon examining the results with a fixed schedule, we see that interaction ratios to restore agents' performance vary across tasks. Therefore, we analyze an adaptive interaction schedule based on the insights of \gls{ood} states causing degenerate performance. We calculate a ratio by dividing the world model loss on evaluation trajectories by the loss on trajectories in the replay buffer. This ratio measures the novelty of the trajectories visited by the current learned policy compared to those seen during training and enables a single threshold for adding self-generated data across tasks.

We set the threshold for the \gls{ood} ratio to 1.35 (see the ablation study in \appref{subsec:ablation}) and inspect it every 5K environment steps over 4 evaluation episodes. 
If the \gls{ood} ratio exceeds this threshold, the \Passive agent collects
2K-step transitions from the environment using its learned policy, denoted as \text{Passive+Auto interact} (refer to \appref{subsec:pseudocode} for the agent's pseudocode).
As shown in \figref{interact}, this strategy fine-tunes self-generated data injection based on task demands, achieving similar performance with less data (5.67\% across 31 tasks) compared to an agent that regularly adds 10\% self-generated data. 
The inspection frequency can be reduced to lower evaluation costs. For more results, see \appref{app:complete-selfgene}. 
This strategy involves online evaluations. A complete offline evaluation would be desirable, but is outside the scope of this paper. We hope to inspire research in this direction. 

\section{Related Work}\label{sec:related}\vspace{-.7em}
\textbf{Performance Degradation in Offline Model-based Agents\quad} 
Performance degradation of offline agents is a known  phenomenon in \gls{mbrl}~\citep{he2023surveyoffline} and is mainly attributed to two factors:

\textbf{(1)~The distribution mismatch between training data and the states visited by the learned policy}~\citep{MOReL,offlineadapt,mopo,cang2021behavioral,Agnostic}.
These inaccuracies in the world model within unseen regions are then exacerbated by compounding errors in multi-step predictions~\citep{comperror,trust}.
These accumulated errors in the model-based imagination process based on \gls{ood} queries can mislead both policy training~\citep{wang2023coplanner} and planning by overestimation in critics~\citep{critics}, ultimately resulting in a performance drop.

\textbf{(2)~The inability of offline agents to self-correct through active data collection} \citep{he2023surveyoffline,cang2021behavioral,mopo}.
Prior works on offline agents~\citep{ostrovski2021difficulty,tang2024understandinggap,emedom2023knowledge,lin2024the} have shown that utilizing data from interactions with the environment introduces a corrective feedback loop~\citep{activecorrect}, allowing the agent to learn from its own mistakes and consequently improve its task performance.

Building on existing studies, we explore phenomena across various tasks and domains in model-based RL using DreamerV3.
Additionally, we investigate the conditions (\eg the nature and quality of the dataset) that exacerbate distribution mismatches and model inaccuracies.

\textbf{Remedies to Support Offline Training\quad} %\label{sec:rel-rem}
To address performance degradation in offline model-based agents, many studies add conservatism to their algorithms. 
One method is to include an uncertainty penalty in the reward function to deter the agent from exploring new states~\citep{MOReL,mopo,combo,wang2023coplanner}, while another employs trust-region updates to maintain the learned policy's proximity to the data collection policy~\citep{matsushima2021deploy}.
RAMBO~\citep{rambo} trains an adversarial environment model that generates pessimistic transitions for \gls{ood} state-action pairs, reducing the value function in uncertain regions. In contrast, MAPLE~\citep{offlineadapt} enables adaptive agent behavior in \gls{ood} regions during deployment, using a context-aware policy based on meta-learning techniques.

While these methods provide insights on mitigating performance degradation in offline \gls{mbrl}, few address which type of data best facilitates offline training.
In model-free RL, studies suggest adding self-generated data~\citep{ostrovski2021difficulty,lee2021offlinetoonline} and emphasize the importance of diversity and exploration~\citep{mediratta2024gengap,suau2023bad,kanitscheider2021multitask,geneoffline,BYOL}.
We extend these ideas to model-based RL with validation in various tasks and domains.

\section{Conclusions and Discussions}\label{sec:conclusion}

We study the effect of data collection for model-based RL agents for offline learning. 
Through a wide range of experiments across various domains, we show that data collection has a huge impact on performance and find that pure offline methods suffer from degraded performance. The reason is that novel states are visited during evaluation. 
This tendency to visit \gls{ood} states arises from the lack of self-correction in offline training on data with limited state-space coverage. The resulting mismatch between imagined and real rollouts misleads policy training and drives agents toward failure.
From a data perspective, we identify that training on partially exploratory data collected using a mixed task-exploration reward function is effective in mitigating performance degradation.
Importantly, training offline solely on expert data exacerbates performance degradation compared to a typical mixed dataset due to severe \gls{ood} issues.
Additionally, our experiments show that adding as little as 10\% self-generated data at regular intervals can significantly enhance the performance of \Passive agents.
When we allow the Passive agent to adaptively interact based on its world model loss as a proxy measure of \gls{ood} state visitation, we observe a significant performance improvement while minimizing the need for additional interaction data. 
However, our method still requires evaluation rollouts. An offline measure would be desirable and is left for future research.

Overall, we highlight the importance of sufficient state-space coverage in the training data to train a robust model-based agent, which can be achieved either by an explorative offline dataset or by enabling the agent to learn from its own mistakes.
As efforts to collect large-scale real-world data for robotics are increasing, the question arises: What is the best way to collect data to facilitate robust agent training?
As model-based RL shows strong task performance and promises efficient fine-tuning and good transfer capabilities for new tasks, we suggest that dataset collection should incorporate exploration data. 
We plan to extend our experiments to other RL methods and real-world scenarios to identify optimal data collection strategies.
We believe that our insights can help design a data-efficient fine-tuning method for robotics foundation models. 
This will help develop more resilient and adaptable agents capable of performing reliably in complex environments.

\section*{Appendix}

\subsubsection*{Acknowledgments} \label{sec:ack}

Funded/Co-funded by the European Union (ERC, REAL-RL, 101045454). Views and opinions expressed are, however, those of the author(s) only and do not necessarily reflect those of the European Union or the European Research Council. Neither the European Union nor the granting authority can be held responsible for them.
This work was supported by the Volkswagen Stiftung (No 98 571) and by the German Federal Ministry of Education and Research (BMBF): Tübingen AI Center, FKZ: 01IS18039A. 
The authors thank the International Max Planck Research School for Intelligent Systems (IMPRS-IS) for supporting CS.
JF is supported by the Max Planck ETH Center for Learning Systems.
Georg Martius is a member of the Machine Learning Cluster of Excellence, funded by the Deutsche Forschungsgemeinschaft (DFG, German Research Foundation) under Germany’s Excellence Strategy – EXC number 2064/1 – Project number 390727645.

%%%%%%%%%%%%%%%%%%%%%%%%%%%%%%%%%%%%%%%%%%%%%%%%%%%%%%%%%%%%%%%%
%% Bibliography
%%%%%%%%%%%%%%%%%%%%%%%%%%%%%%%%%%%%%%%%%%%%%%%%%%%%%%%%%%%%%%%%
\bibliographystyle{rlj}
\bibliography{ref}

%%%%%%%%%%%%%%%%%%%%%%%%%%%%%%%%%%%%%%%%%%%%%%%%%%%%%%%%%%%%%%%%
% AUTHOR: If your paper has no supplementary materials, you may 
%         comment out the line below, which creates the title for
%         the supplementary materials.
%%%%%%%%%%%%%%%%%%%%%%%%%%%%%%%%%%%%%%%%%%%%%%%%%%%%%%%%%%%%%%%%
\beginSupplementaryMaterials

%Do this for an archiv submission where you want all in one file. (comment out the cross referencing myexternal... above)
%\input{suppl_content.tex}
% Figures, Tables and Equations will have S in the name
\renewcommand{\thetable}{S\arabic{table}}
\renewcommand{\thefigure}{S\arabic{figure}}
\renewcommand{\theequation}{S\arabic{equation}}
\setcounter{table}{0}
\setcounter{figure}{0}
\setcounter{equation}{0}
\appendix

\section{Implementation Details}\label{sec:hyperparam}
\subsection{Runtime Overview}
Our experiments comprised approximately 2000 runs, totaling 20000 GPU hours. 
Each run took between 8 and 15 hours, depending on the specific task. All experiments were conducted using NVIDIA RTX 4090 or A100 GPUs.

\subsection{Model Hyperparameters}
For all experiments, we use the same model size $S$, defined in~\citet{hafner2023mastering}.
Each agent, which consists of a world model, an actor network, and a critic network, has a total of 18M optimizable variables.
We follow the default values in~\citet{hafner2023mastering} for the training hyperparameters \eg learning rate and batch size for each component of the agent as well as other hyperparameters.
For more details about DreamerV3, please refer to~\citet{hafner2023mastering}.
\subsection{Environment Hyperparameters}
We list the environment hyperparameters in \tabref{tab:env_param}.
The implementation of the task \emph{Point Mass Maze} is based on~\citet{yarats2022exorl}.
\begin{table}[htb]
\centering
\caption{Environment hyperparameters for each domain}
\begin{tabular}{cccc}
\toprule
\textbf{Hyperparameter}  & \textbf{\gls{dmc}} & \textbf{Metaworld} & \textbf{MinAtar} \\
\midrule
Image Size & [64,64] & [64,64] & [32,32] \\
Action Repeat & 2 & 2 & 1 \\
Episode Truncate & - & - & 2500 \\
Parallel Env Num & 4 & 4 & 4 \\
Train Ratio & 512 & 512 & 512 \\
\bottomrule
\end{tabular}
\label{tab:env_param}
\end{table}
\subsection{Environment Steps in Offline Agents}\label{subsec:explain-envsteps}
Tracking performance metrics relative to environment steps during online training is standard practice in the RL community. This methodology is also applied in the analysis of the offline \Tandem agent in~\citet{ostrovski2021difficulty}, which closely mirrors the behavior of its \Active counterpart.

However, the \Passive agent—by definition—does not interact with the environment and thus cannot influence environment steps. This poses a challenge for directly comparing its performance with that of the \Active and \Tandem agents. To ensure comparability across training procedures, we allow the \Passive agent to interact with the environment during training in the same manner as an online agent, but without adding the resulting interaction data into its replay buffer. This setup enables the \Passive agent to remain trained solely on an offline dataset while allowing performance comparisons based on environment steps, with only minimal code changes required.

\subsection{Pseudocode of methods}\label{subsec:pseudocode}
We add the pseudocode of the \Active, \Passive, and \Tandem agents (in \algref{alg:agents}), the variation of \Passive and \Tandem agents (in \algref{alg:agents_cont}) used in decoupled analysis (\secref{subsubsec:why-ood}), as well as the second remedy (in \algref{alg:agents_interact}) for better clarity.
Below, the training of the world model $M$ includes training all components in \eqref{eq:wm}, while training $\pi$ includes all components in \eqref{eq:ac}.

\begin{algorithm}[H]
\caption{Learning agents}
\label{alg:agents}
\vspace{-1em}
\begin{multicols}{3}
% Active Agent
\centering\textbf{\textcolor{ActiveColor}{Active Agent}}
\begin{algorithmic}[1]
\State \textbf{Initialize:} Replay buffer $\mathcal{B}$ = a few random episodes.
\State  World model $M$ + Policy $\pi$ by seed \textcolor{ActiveColor}{$S_{A}$}.
\For{each step $i$}
    \State Sample $ \textcolor{ActiveColor}{\mathcal{D}^i_A} \sim \mathcal{B}$
    \State Update $M$ using $\mathcal{D}^i_A$
    \State Train $\pi$ in the imagination of $M$
    \State Execute $\pi$ in the env to expand $\mathcal{B}$
\EndFor
\State \textbf{Return:} Final \textcolor{ActiveColor}{$\mathcal{B}_{A}$}, $\pi$
\end{algorithmic}
\columnbreak
% Passive Agent
\centering\textbf{\textcolor{PassiveColor}{Passive Agent}}
\begin{algorithmic}[1]
\State \textbf{Initialize:} Replay buffer $\mathcal{B}$ = final \textcolor{ActiveColor}{$\mathcal{B}_{A}$}.
\State World model $M$ + Policy $\pi$ by seed \textcolor{PassiveColor}{$S_{P}$}.
\For{each step $i$}
    \State Sample $\mathcal{D}^i_P \sim \mathcal{B}$
    \State Update $M$ using $\mathcal{D}^i_P$
    \State Train $\pi$ in the imagination of $M$
    \Statex \textcolor{PassiveColor}{-}
    \Statex \textcolor{PassiveColor}{-}
    \vspace{0.35em}
\EndFor
\State \textbf{Return:} $\pi$
\end{algorithmic}
\columnbreak
% Tandem Agent
\centering\textbf{\textcolor{TandemColor}{Tandem Agent}}
\begin{algorithmic}[1]
\State \textbf{Initialize:} Replay buffer $\mathcal{B}$ =  final \textcolor{ActiveColor}{$\mathcal{B}_{A}$}.
\State World model $M$ + Policy $\pi$ by seed \textcolor{TandemColor}{$S_{T}$}.
\For{each step $i$}
    \State \textcolor{TandemColor}{Copy} $\textcolor{TandemColor}{\mathcal{D}^i_T =} \textcolor{ActiveColor}{\mathcal{D}^i_A}$
    \State Update $M$ using $\mathcal{D}^i_T$
    \State Train $\pi$ in the imagination of $M$
    \Statex \textcolor{TandemColor}{-}
    \Statex \textcolor{TandemColor}{-}
    \vspace{0.25em}
\EndFor
\State \textbf{Return:} $\pi$
\end{algorithmic}
\end{multicols}
\vspace{-0.7em}
\end{algorithm}

\begin{algorithm}[H]
\caption{Learning agents in decoupled analysis}
\label{alg:agents_cont}
\vspace{-1em}
\begin{multicols}{3}
% Active Agent
\centering\textbf{\textcolor{ActiveColor}{Active Agent}}
\begin{algorithmic}[1]
\State \textbf{Initialize:} Replay buffer $\mathcal{B}$ = a few random episodes.
\State  World model $M$ + Policy $\pi$ by seed \textcolor{ActiveColor}{$S_{A}$}.
\Statex \textcolor{ActiveColor}{-}
\For{each step $i$}
    \State Sample $ \textcolor{ActiveColor}{\mathcal{D}^i_A} \sim \mathcal{B}$
    \State Update $\textcolor{ActiveColor}{M^i_A}$ using $\mathcal{D}^i_A$
    \State Train $\pi$ in the imagination of $M$
    \State Execute $\pi$ in the env to expand $\mathcal{B}$
\EndFor
\State \textbf{Return:} Final \textcolor{ActiveColor}{$\mathcal{B}_{A}$}, \textcolor{ActiveColor}{$M_A$}, $\pi$
\end{algorithmic}
\columnbreak
% Passive_sameWM(frozen) Agent
\centering\textbf{\textcolor{p}{Passive\_sameWM (frozen)}}
\begin{algorithmic}[1]
\State \textbf{Initialize:} Replay buffer  $\mathcal{B}$ = final \textcolor{ActiveColor}{$\mathcal{B}_{A}$}.
\State World model $M$ = final \textcolor{ActiveColor}{$M_A$}, only Policy $\pi$ by seed \textcolor{p}{$S_{P}$}.
\For{each step $i$}
    \State Sample $\mathcal{D}^i_P \sim \mathcal{B}$
    \Statex \textcolor{p}{-}
    \State Train $\pi$ in the imagination of $M$
    \Statex \textcolor{p}{-}
    \Statex \textcolor{p}{-}
    \vspace{0.35em}
\EndFor
\State \textbf{Return:} $\pi$
\end{algorithmic}
\columnbreak
% Tandem_sameWM Agent
\centering\textbf{\textcolor{y}{Tandem\_sameWM}}
\begin{algorithmic}[1]
\State \textbf{Initialize:} Replay buffer $\mathcal{B}$  = final \textcolor{ActiveColor}{$\mathcal{B}_{A}$}.
\State only Policy $\pi$ by seed \textcolor{y}{$S_{T}$}.
\Statex \textcolor{y}{-}
\Statex \textcolor{y}{-}
\For{each step $i$}
    \State \textcolor{y}{Copy} $\textcolor{y}{\mathcal{D}^i_T =} \textcolor{ActiveColor}{\mathcal{D}^i_A}$
    \State \textcolor{y}{Copy} $\textcolor{y}{M =} \textcolor{ActiveColor}{M^i_A}$
    \State Train $\pi$ in the imagination of $M$
    \Statex \textcolor{y}{-}
    \Statex \textcolor{y}{-}
    \vspace{0.25em}
\EndFor
\State \textbf{Return:} $\pi$
\end{algorithmic}
\end{multicols}
\vspace{-0.7em}
\end{algorithm}

\begin{algorithm}[H]
\caption{Passive agents adding additional self-generated data ($\mathrm{K}$ denotes thousand)}
\label{alg:agents_interact}
\vspace{-1em}
\begin{multicols}{3}
% Passive Agent
\centering\textbf{\textcolor{PassiveColor}{Passive Agent}}
\begin{algorithmic}[1]
\State \textbf{Initialize:} Replay buffer $\mathcal{B}$ = final $\mathcal{B}_{A}$.
\State World model $M$ + Policy $\pi$ by seed $S_{P}$.
\For{each step $i$}
    \State Sample $\mathcal{D}^i \sim \mathcal{B}$
    \State Update $M$ using $\mathcal{D}^i$
    \State Train $\pi$ in the imagination of $M$
    \Statex \textcolor{PassiveColor}{-}
    \Statex \textcolor{PassiveColor}{-}
    \Statex \textcolor{PassiveColor}{-}
    \Statex \textcolor{PassiveColor}{-}
    \vspace{0.25em}
\EndFor
\State \textbf{Return:} $\pi$
\end{algorithmic}
\columnbreak
% Passive Agent
\centering\textbf{\textcolor{y}{Fixed Schedule}}
\begin{algorithmic}[1]
\State \textbf{Initialize:} Replay buffer $\mathcal{B}$ = final $\mathcal{B}_{A}$.
\State World model $M$ + Policy $\pi$ by seed $S_{P}$.
\For{each step $i$}
    \State Sample $\mathcal{D}^i \sim \mathcal{B}$
    \State Update $M$ using $\mathcal{D}^i$
    \State Train $\pi$ in the imagination of $M$
    \If{\textcolor{y}{$i \% N == 0$ }}  \Comment{$N=4\mathrm{K}, 20\mathrm{K}, 200\mathrm{K}$}
        \State Execute $\pi$ in the env to expand $\mathcal{B}$ by $2\mathrm{K}$ step data
    \EndIf 
\EndFor
\State \textbf{Return:} Final $\mathcal{B}$, $\pi$
\end{algorithmic}
\columnbreak
% Passive Agent
\centering\textbf{\textcolor{p}{Adaptive Schedule}}
\begin{algorithmic}[1]
\State \textbf{Initialize:} Replay buffer $\mathcal{B}$ = final $\mathcal{B}_{A}$.
\State World model $M$ + Policy $\pi$ by seed $S_{P}$.
\For{each step $i$}
    \State Sample $\mathcal{D}^i \sim \mathcal{B}$
    \State Update $M$ using $\mathcal{D}^i$
    \State Train $\pi$ in the imagination of $M$
    \If{\textcolor{p}{$i \% 5\mathrm{K} == 0$ and $\mathrm{ood\_ratio}_i > \mathrm{thres.}$}}
        \State Execute $\pi$ in the env to expand $\mathcal{B}$ by $2\mathrm{K}$ step data
    \EndIf 
\EndFor
\State \textbf{Return:} Final $\mathcal{B}$, $\pi$
\end{algorithmic}
\end{multicols}
\vspace{-0.7em}
\end{algorithm}

\subsection{Supplementary of DreamerV3}\label{sec:dreamer-app}
The computation of each component in the world model loss:
\eq{
\mathcal{L_{\mathrm{pred}}}(\phi) & \doteq
-\lnpp(x_t|z_t,h_t)
-\lnpp(r_t|z_t,h_t)
-\lnpp(c_t|z_t,h_t) \\
\mathcal{L_{\mathrm{dyn}}}(\phi) & \doteq
\max\bigl(1,\KL[\sg(\qp(z_t|h_t,x_t)) || \hspace{3.2ex}\pp(\hat z_t|h_t)\hphantom{)}]\bigr) \\
\mathcal{L_{\mathrm{rep}}}(\phi) & \doteq
\max\bigl(1,\KL[\hspace{3.2ex}\qp(z_t|h_t,x_t)\hphantom{)} || \sg(\pp(\hat z_t|h_t))]\bigr)
\label{eq:wm_loss-comps}
}

\subsection{Ablation Studies}\label{subsec:ablation}
\begin{figure}[tb]
   \centering
   \includegraphics[scale=1.0]{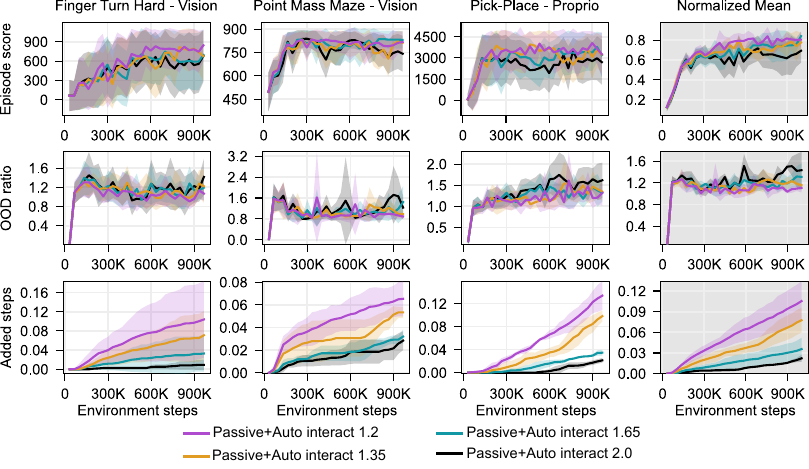}
   \caption{\textbf{Ablation studies on threshold value for adaptive \Passive agents.} We test four threshold values: 2.0, 1.65, 1.35, and 1.2 in three tasks. The last column shows a normalized mean across tasks. The number of added steps in the third row is shown as a percentage of the original replay buffer size.}
   \label{ablation}
   \vspace{-5pt}
\end{figure}

We test different threshold values used in adaptive \Passive agents for autonomously adding self-generated interaction data.
In \figref{interact_all_ood}, we observe that the majority \gls{ood} ratio in \Active agents reaches below 2.0 during training.
Therefore, we begin with an upper bound threshold value of 2.0 and test four values: 2.0, 1.65, 1.35, and 1.2.
It is important to note that this upper bound serves solely as a reference point for initiating the ablation studies and does not imply any dependence of the \gls{ood}\ ratio on the performance of the \Active agent.
In \figref{ablation}, we show that although a lower threshold value (\eg 1.2) could bring more self-generated data (about 10\% average) to the replay buffer, the improvement in performance is not significant compared to other higher values.
However, a high threshold value (\eg 2.0) makes the training process less stable, as shown in the relatively low normalized mean score and an increasing tendency of \gls{ood} ratio from step 800K, compared to lower threshold values.
But generally, the sensitivity of this threshold value to performance is low.
One can set a low threshold value if the training budget allows.
In the main experiments, we choose a middle threshold value of 1.35, which balances the number of added interaction data and stable performance.
Although 1.65 performs similarly with fewer samples in the three test environments, we conservatively adopt 1.35 to ensure greater stability in potentially more challenging settings.

\subsection{Evaluation Episodes Details for Metrics Computation} \label{subsec:eval-details}
During training, each evaluation phase involves rolling out the agent's policy for a total of 4 episodes. 
The episode score (described in \secref{sec:met}) is computed on the entire episode trajectory. 
For other metrics—such as world model loss—we collect up to 500 steps per episode using a first-in-first-out (FIFO) buffer. 
The world model loss is computed over the collected steps and averaged to obtain a mean value for each episode. 
Additional metrics, such as the value function (described in \secref{subsec:add-mets}), are computed at the first step of the buffer. 
All metric values are then averaged across the 4 evaluation episodes.

\subsection{Additional Metrics} \label{subsec:add-mets}
\textbf{Policy input reconstruction loss\quad}
We train an autoencoder functioning as an \gls{ood} detector for the policy inputs. The autoencoder is optimized to minimize the negative log-likelihood \eqrefp{eq:policy_recon_loss} to reconstruct the policy input.
Novel policy inputs, that may compromise the quality of output actions, can be detected using the \gls{mse} reconstruction loss.
A higher \gls{mse} indicates that the input is likely novel or anomalous, suggesting the input differs significantly from the training distribution and could lead to an unreliable policy action.
\eq{
\mathcal{L_{\mathrm{recon}}}(\phi) & \doteq
-\lnpp(z_t,h_t \mid 
\text{encoder}(z_t,h_t))
\label{eq:policy_recon_loss}
}
\textbf{Value function\quad}
The expected discounted return—the cumulative sum of future rewards, as shown in \eqref{mbrl-goal}.

The additional metrics are calculated as follows unless specified otherwise: (1)~Every 5K environment steps, we roll out the agent's policy for a total of 4 episodes. (2)~We compute the policy input reconstruction loss across the 4 episodes. For the value function, we calculate it at the initial state of the collected trajectory per episode and then average these values across the 4 episodes. The implementation details can be found in \secref{subsec:eval-details}.

\section{Result Analyses}\label{sec:results}
\subsection{Discrepancy between Imagination and Real Rollouts}\label{sec:discrepancy}
\begin{figure}[tb]
   \centering
   \includegraphics[scale=0.9]{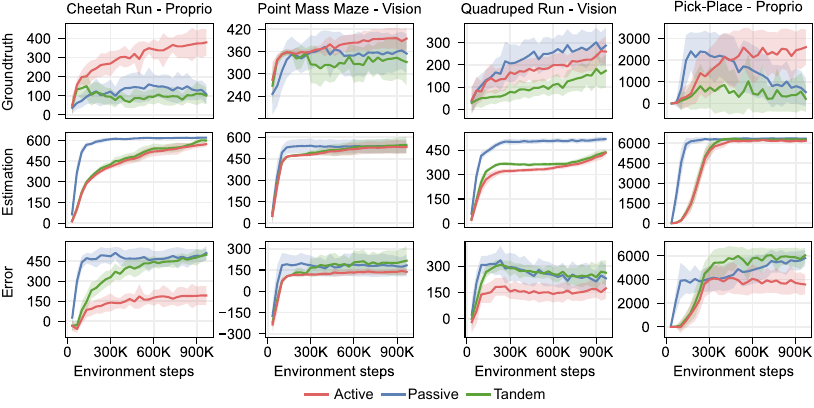}
   \vspace{-.7em}
   \caption{\textbf{Value function estimation of each agent in 4 selected tasks.} The value function $V(s)$ is calculated on the initial state of each agent's collected trajectory, which should reflect the actual discounted rewards accumulated across the trajectory. The ground truth value is computed using Monte Carlo estimation from one sample trajectory. The error is computed by subtracting the ground truth value from the estimated value.}
   \label{discrepancy}
   \vspace{-5pt}
\end{figure}
As outlined in \secref{sec:preli}, the agent’s policy utilizes an actor-critic framework, with the critic predicting the value function $V(s)$ for each given state.
Since the critic is trained in the imagination of the world model and will subsequently be used to train the actor, it is essential that its value estimates accurately reflect the agent’s real rollout conditions. 
If the actual rollout performs poorly, a correct low-value estimate from the critic can guide the actor’s updates in a direction that improves performance.
However, in \figref{discrepancy}, we show that both \Passive and \Tandem agents consistently wrongly predict their value functions, assigning high values even when their actual trajectories yield low rewards. 
Throughout training, the value function estimation error for these offline agents remains significantly higher than that of the \Active agent, showing consistent statistical differences across time scales.
This finding highlights that, without the self-correction mechanism, offline agents develop a pronounced mismatch between imagined and real rollouts, which is reflected in the discrepancy between estimated and ground-truth returns.
This misalignment can lead to suboptimal actor updates in many tasks, ultimately resulting in unstable or degraded performance.
Nevertheless, in certain tasks such as \emph{Quadruped Run - Vision}, the \Passive agent appears less affected by its value overestimation, suggesting that other task-specific factors may mitigate the impact of inaccurate value estimation.

Moreover, in some tasks such as \emph{Pick-Place - Proprio}, the magnitude of the error is even larger than that of the ground truth, even for the \Active agent.
This observation raises the question of whether the standard practice of using value function error is the most ideal metric for quantifying the imagination gap.
Since value functions are trained via bootstrapping, they are inherently biased and unlikely to converge exactly to Monte Carlo returns.
A potentially better alternative could be to compare the average predicted reward (used in boostrapping to train the critic) with the actual reward collected during evaluation.
We leave a detailed investigation of such alternatives to future work.
The full results including more tasks can be found in \secref{app:complete-valuefn}.

\subsection{Per-step Analysis of Performance Degradation}\label{sec:step-analysis}
\begin{figure}[tb]
   \centering
   \includegraphics[scale=1.0]{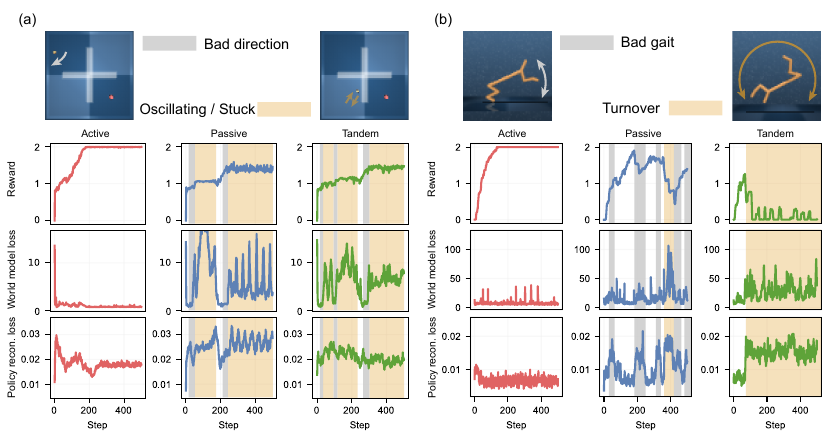}
   \caption{\textbf{Stepwise analysis within a single test episode} of the Point Mass Maze - Vision and Cheetah Run - Vision tasks from \gls{dmc}. The plots show the progression of reward, world model loss, and policy input reconstruction loss at each step as the agent executes actions given by its own policy. Timesteps, where agents exhibit abnormal behavior, are highlighted with yellow and grey regions. Each episode consists of 500 steps, with the environments initialized identically across agents. The agents are the fully trained version after 1M environment steps.}
   \label{step_analysis}
   \vspace{-5pt}
\end{figure}

\textbf{Compromised policy causes the agent to drift into novel states from familiar regions.}
This is particularly evident in task (a) \emph{Point Mass Maze - Vision} of \figref{step_analysis}.
In the timesteps marked by the grey regions—just before entering novel states—the agent exhibits low world model loss, yet its movement direction deviates from the typical task-solving trajectory.
This suggests that the policy is already compromised, even before encountering novel states, matching the argument in~\secref{sec:toy}.
However, in the locomotion environment (b) \emph{Cheetah Run - Vision}, this is less observable, which could be attributed to the nature of the task—where it is more difficult to visually identify subtle changes in gait compared to tracking the trajectory of a ball.

\textbf{Catastrophic cycle: novel states disrupt world model and policy output during evaluation.}
After the agent enters into novel states, the world model will output inaccurate estimations and latent embeddings.
Since the policy network relies on these inaccurate latent states as input, this can start the catastrophic cycle where each compromised action leads to further novel states and additional inaccuracies until the episode ends or the agent accidentally re-enters into a familiar state.
In \figref{step_analysis}, we provide for two test times trajectories the reward, world model loss, and policy reconstruction loss across two tasks. A low task reward is typically accompanied by a high world model loss.
A high world model loss can correlate with a high policy input reconstruction loss, suggesting that the policy is unfamiliar with such inputs and produces compromised actions—an indication of the catastrophic cycle.
In both tasks, the catastrophic cycle is evident in the significant failure periods (\eg oscillation in \emph{Point Mass Maze - Vision} or turnover in \emph{Cheetah Run - Vision}).
In addition, the accidental re-entering into a familiar state can also be observed between the neighboring failure periods.
However, this accidental re-entering cannot save the agent since its policy is already compromised due to the lack of self-correction mechanism and cannot consistently output a reliable action in familiar states.

\textbf{World model can sometimes hallucinate and mislead policy in novel states.}
\begin{figure}[tb]
   \centering
   \includegraphics[scale=1.0]{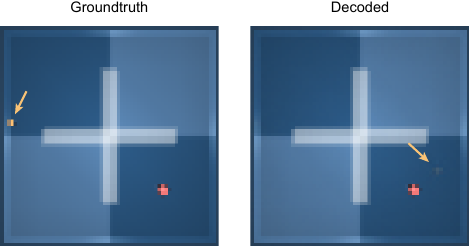}
   \caption{\textbf{World model misinterprets the novel states.} In the decoded image (step 84 in \figref{step_analysis}) from the world model of the \Passive agent in task Point Mass Maze - Vision, the ball vaguely appears near the goal position while in the ground truth observation, it is actually in a novel region to the world model.}
   \label{wrong_mapping}
   \vspace{-5pt}
\end{figure}
We observe unexpected instances where the world model hallucinates, as shown in \figref{wrong_mapping}.
The decoded image by the world model shows the agent has already reached a position near the goal, while, in fact, it is still far away from the target.
It indicates that the world model can hallucinate in the novel states and produce an incorrect mapping of the latent state, misleading the policy to output inadequate actions.

\textbf{Sometimes, policy input reconstruction loss fails to reflect performance degradation.}
This is observed in the \Tandem agent in \emph{Point Mass Maze - Vision} in \figref{step_analysis}.
In novel states, a high world model loss may still result in a relatively low policy input reconstruction loss, possibly due to erroneous latent state mapping produced by the world model.

\subsection{Detailed Results of Considerations in Practical Applications}\label{sec:ood-app-detail}
\begin{figure}[tb]
   \centering
   \includegraphics[scale=1.0]{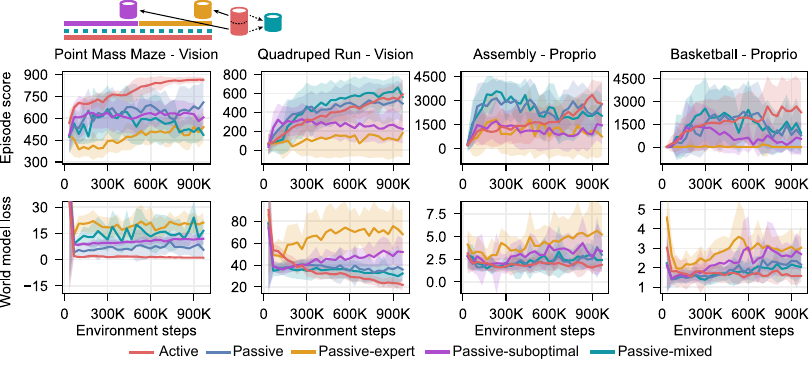}
   \caption{\textbf{Performance comparison when training \Passive agents on different halves of the replay buffer from the \Active.} We split the replay buffer (red bucket) at the 500K environment steps, as shown in the schematic illustration on the Point Mass Maze - Vision. The first half (purple bucket) represents the suboptimal data, while the second half (yellow bucket) mainly contains high-reward expert data. Therefore, Passive-expert, Passive-suboptimal, and Passive-mixed have a halved replay buffer compared to the normal \Passive agent. The replay buffer of the mixed agent (turquoise bucket) is uniformly sampled from the whole replay buffer.}
   \label{split}
   \vspace{-5pt}
\end{figure}

\paragraph{Advantage of training agents offline}
Although the performance degradation caused by the \gls{ood} issue is prominent in \Passive agents, they show potential for faster convergence and more efficient training, as seen in tasks like \emph{Quadruped Run - Vision} and \emph{Pick-Place - Proprio} in \figref{episode_score_wmloss_4tasks}.
This is because \Passive agents have access to high-quality trajectories from the beginning, while \Active agents must wait until later in training to encounter those trajectories.
We validate this hypothesis in \figref{split}, where \Passive agents trained on suboptimal data generally perform worse than those trained on mixed data.
It indicates that mixing expert trajectories into suboptimal data helps the performance, which matches the case between the \Active (suboptimal data) vs. \Passive (mixed data) agent in the early training stage.
Therefore, addressing the \gls{ood} issue in \Passive agents is crucial, as solving it could unlock the potential for highly efficient agent training.
However, we do not observe such advantages in \Tandem agents.

\paragraph{Different model initialization}
\begin{figure}[tb]
   \centering
   \includegraphics[scale=1.0]{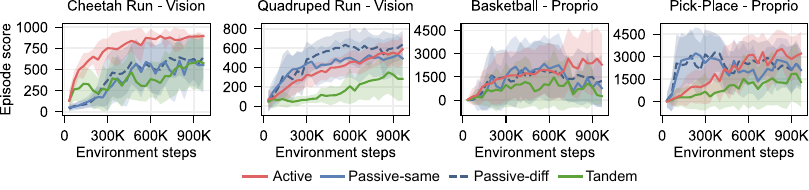}
   \caption{\textbf{Model initialization matters not in \Passive agents.} Performance comparison when initializing the world model and policy network of \Passive agents with the same and different seed \wrt the \Active agents.}
   \label{seed}
   \vspace{-5pt}
\end{figure}
In this section, we answer the question whether the model initialization affects the performance degradation.
In particular, if we initialize the world model and policy network of a \Passive agent using the same seed as the \Active one, will the performance differ from the independently initialized \Passive agent?
In \figref{seed}, we show that no significant difference in the task performance can be observed with initialization seeds among \Passive agents.
We also investigate the sensitivity of task performance to the initialization of weights in model networks of Tandem agents.
By mixing weights of the identically initialized networks as the \Active and those of an independent initialization with different ratios $\alpha$, it allows us to observe whether a tiny difference in the initialization will cause a big difference in task performance.
\eq{
w & \doteq
(1-\alpha) \cdot w_\mathrm{\Active} + \alpha \cdot  w_\mathrm{\Tandem}
\label{eq:mixweight}
}

\begin{wrapfigure}[13]{r}{.35\textwidth}
    \vspace{-1.7em}
   \centering
   \includegraphics[scale=1.0]{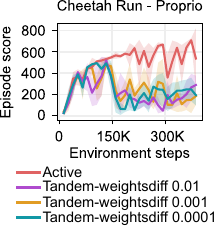}
   \vspace{-.8em}
   \caption{\textbf{Performance comparison of the world model and policy network of \Tandem agents initialized with mixed weights.} Results shown for different $\alpha$ values (indicated in run name) as defined in \eqref{eq:mixweight}. Results for one seed.
   %The value of $\alpha$ is indicated in the agent's name. The experiment is run with a single seed.
   }

   \label{mixweight}
   \vspace{-10pt}
\end{wrapfigure}

In \figref{mixweight}, we observe that even a small deviation from the weights of the \Active agent eventually causes a large difference in task performance when training on the identical sequence of training batches each training step.

\paragraph{World model overfitting on expert dataset}
Another popular practice to facilitate training a capable agent is to train the agent on an expert dataset~\citep{kumar2022offline}.
However, in \figref{split}, we find that training on expert data leads to an even worse performance degradation in \Passive agents.
It is also indicated by the high world model loss with a growing tendency.
However, according to the performance of Passive-mixed agents, mixing expert data with suboptimal trajectories can help mitigate this issue.
The expert dataset primarily consists of monotonic task-solving trajectories, which implies extremely limited state-space coverage.
Incorporating suboptimal data expands this coverage during training.
It can improve the generalization capability of the world model and reduce the \gls{ood} risk during policy rollouts in evaluation.
This highlights the importance of broad state-space coverage during training and the need to include exploration-equivalent data to ensure a capable agent.
This finding matches results from previous research~\citep {gulcehre2021regularized,mediratta2024gengap,suau2023bad}.

\paragraph{World model overfitting on low-dimensional inputs}
In the \emph{Basketball - Proprio} and \emph{Pick-Place - Proprio} tasks, the performance of the \Passive agent declines as the world model loss increases in the second half of the training process.
A similar issue is observed in proprioceptive versions of \gls{dmc} tasks in \appref{app:complete-task}.
It indicates that the world model begins to overfit on the fixed data distribution in the replay buffer, given that the \Passive agent is not allowed to add its own interaction data and cannot change the data distribution progressively in the same way as the \Active agent.
This tendency is more pronounced in the proprioceptive version, likely because the lower input dimension of the world model, compared to image-based observations, makes it more susceptible to overfitting when combined with high model capacity.

\subsection{Further Insights on Self-Correction Mechanism}\label{sec:more-self-correction}

Here, we present additional perspectives on the self-correction mechanism. 
Further experiments are expected to strengthen the arguments, and we leave them to future work.

\paragraph{Copying data from an online agent is not self-correction}
In \figref{task_all_score}, we observe that the \Tandem agents generally suffer from performance degradation \wrt the \Active agents.
They tend to visit \gls{ood} states more often because the policy/actor training suffers from misleading critic prediction as shown in~\figref{discrepancy_all}.
Although the \Tandem agents share the exact same training data stream (for the world model training) as the \Active agents, they do not benefit from the self-correction in the online agents.
Therefore, simply copying the step-wise training data from an online agent is not sufficient for self-correction.
Since the \Tandem agent begin with different weights and therefore a different policy from the one in the \Active agent, it might need different set of training data for effective self-correction.

\paragraph{Indirect self-correction via shared world models and the need for ongoing self-correction}
From the results in~\figref{wmsame} as discussed in~\secref{subsec:deep-dive}, we identify that discrepancies in both world model and policy contribute to performance degradation, with their relative impacts depending on the task.
However, from a different perspective, this can be understood as: by using the same world model as the online agent, the Tandem\_sameWM agents can improve their performance \wrt the original \Tandem agents to varying degrees.
This suggests that the self-correction in an online-learning agent can indirectly benefit a different policy in another agent via a shared world model.
In other words, even if the task policy itself does not engage in the process of self-correction, an online-trained world model—being generally more accurate than its offline-trained counterpart—can still improve task performance in a tandem setting.
This could possibly be viewed as an indirect self-correction effect.
By contrast, the Passive\_sameWM(frozen) agent shows no such benefit, likely because its world model is fixed during training, with no online feedback loop.
This may also indicate the importance of an ongoing self-correction process, as in an online-learning agent.
If the self-correction stops, then the agent may begin to suffer from policy exploiting the generalization error in the world model, ultimately harming performance.
Additional experiments \eg, stopping the online feedback process in the middle of training, could further validate this argument, and we leave such investigations to future work.

\paragraph{Self-correction in exploration setting}

In \secref{sec:exploration}, we compare online (\Active) and offline (\Passive, \Tandem) agents in a pure exploration setting, where the \Active agent collects interaction data via an online-trained exploration policy while training a task policy in parallel.
Since data collection is driven by the exploration policy rather than the task policy, one might question whether the \Active agent provides any self-correction for the task policy, and whether it serves as a fair baseline for measuring offline degradation.
However, as noted earlier, the task policy may still benefit from self-correction indirectly via the shared world model with the exploration policy.
We therefore retain it as a baseline, and additionally compare to the online task-oriented agent to support the argument that exploration data can counteract the lack of self-correction and help mitigate the \gls{ood} challenge during evaluation in offline agents.

\subsection{Complete Results}\label{app:complete}
\subsubsection{Results of Agents with Different Exploration Bonus}\label{app:complete-expl-bonus}
\begin{figure}[tb]
   \centering
   \includegraphics[scale=1.0]{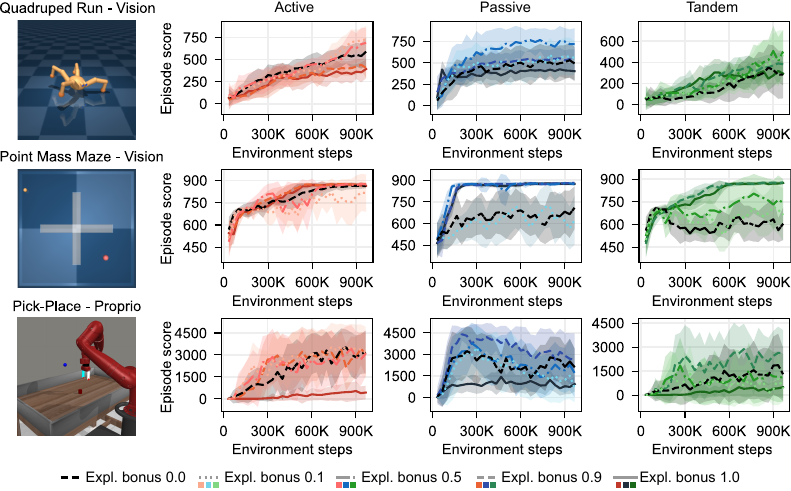}
   \caption{\textbf{Different task has different optimal exploration bonus values.} Performance comparison when assigning different exploration bonuses $w_\mathrm{expl}$ in the reward function. The black dashed lines represent pure task-oriented policy without any exploration bonus.}
   \label{explore_bonus_full}
   \vspace{-5pt}
\end{figure}

In \figref{explore_bonus_full}, we show all three analyzed tasks with comparison among different exploration bonus values.
The optimal exploration bonus $w_\mathrm{expl}$ is 0.5 for task \emph{Quadruped Run - Vision}, 0.9 for tasks \emph{Point Mass Maze - Vision} and \emph{Pick-Place - Proprio}.

\subsubsection{Results of Task-oriented Agents}\label{app:complete-task}
\begin{figure}[tb]
   \centering
   \includegraphics[scale=0.95]{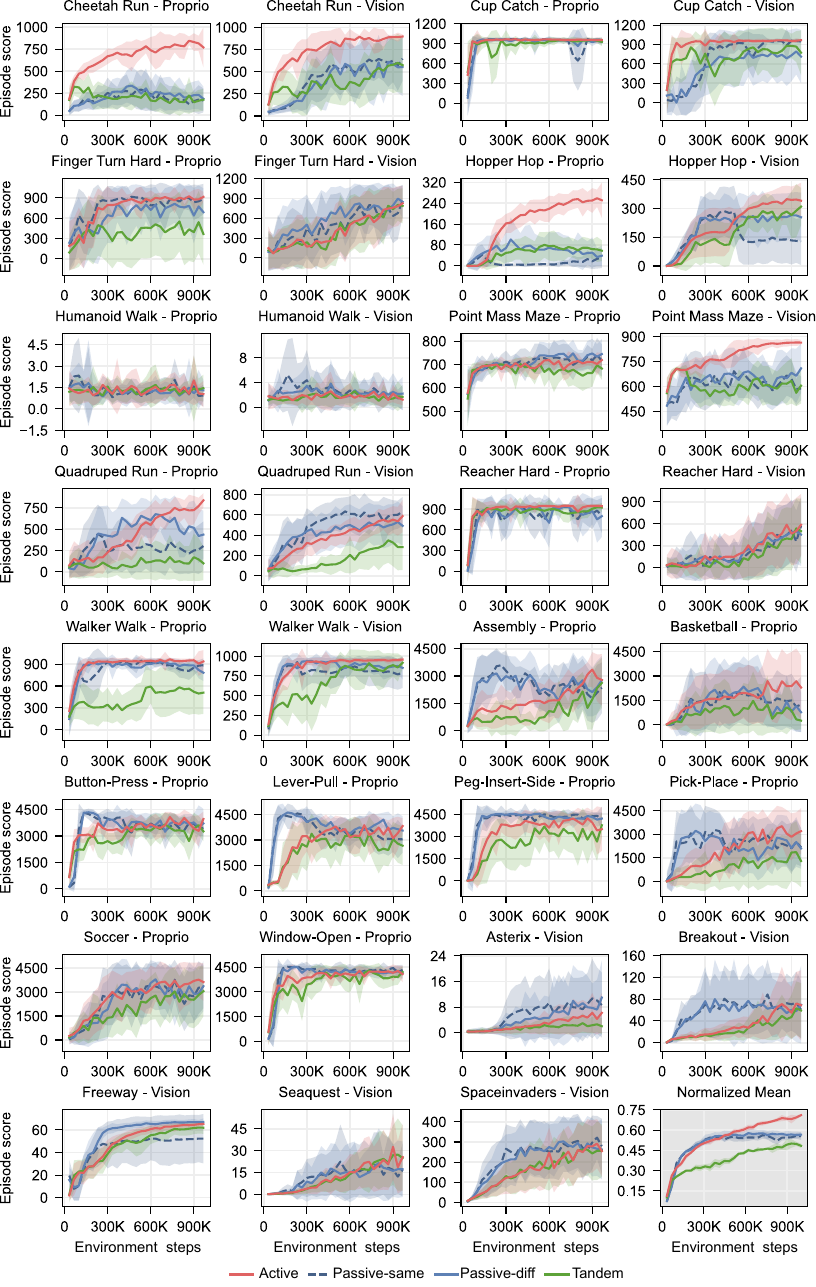}
   \caption{\textbf{Episode score of 31 tasks.} The first 18 tasks are from \gls{dmc}, the subsequent 8 tasks are from Metaworld, and the last 5 are from the MinAtar domain. We also output a normalized mean score across tasks. The Passive-same is \Passive agents initialized identically as the \Active agents while Passive-diff is independently initialized.}
   \label{task_all_score}
   \vspace{-5pt}
\end{figure}

\begin{figure}[tb]
   \centering
   \includegraphics[scale=0.95]{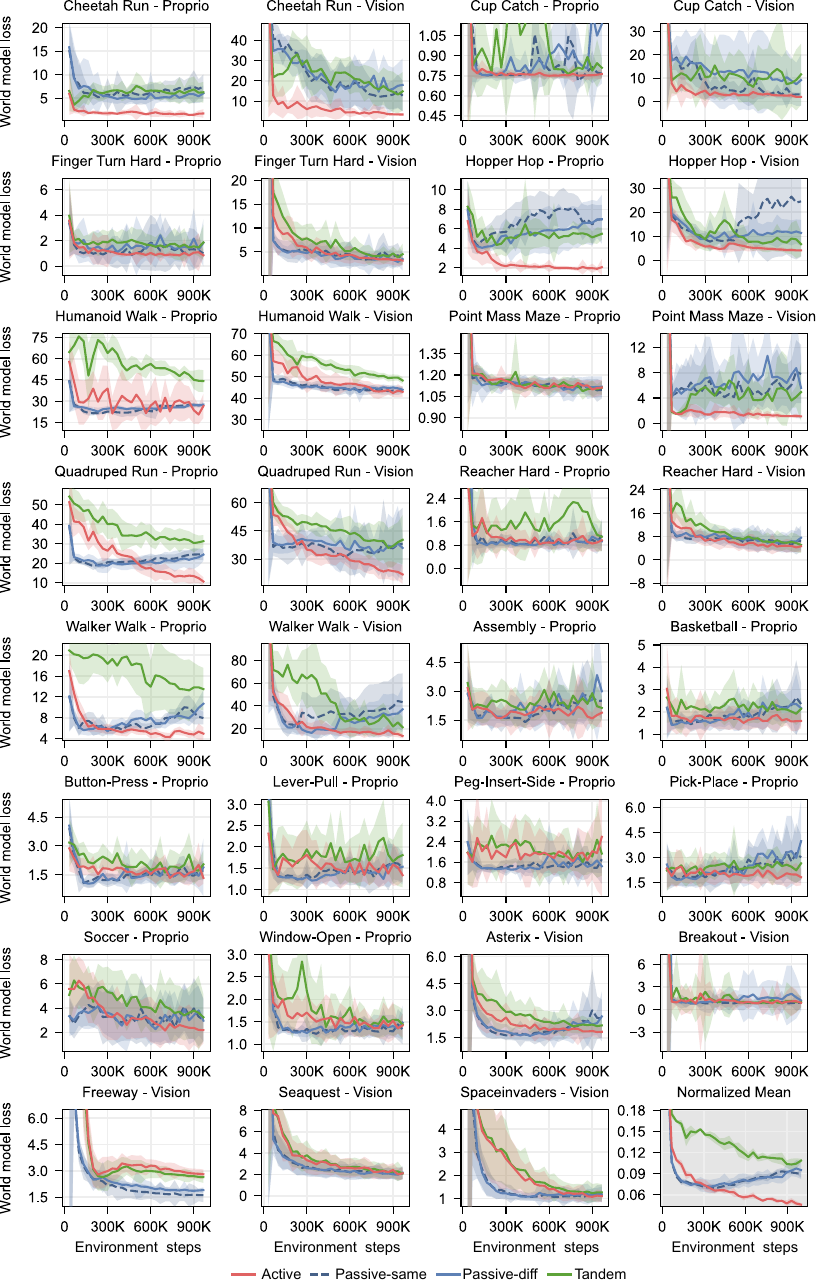}
   \caption{\textbf{World model loss of 31 tasks.} In the last subplot, we show an additional normalized mean result across tasks.}
   \label{task_all_wm}
   \vspace{-5pt}
\end{figure}

In \figref{task_all_score} and \figref{task_all_wm}, we show the complete results in 31 tasks corresponding to the discussion in \secref{sec:val-tasks} and \secref{subsec:deep-dive}.
The \Passive agent initialized using the same seed for the world model and policy network as the \Active agent is marked with a suffix ``-same'', while the different model initialization is marked with ``-diff''.

\subsubsection{Results of Adding Self-generated Data}\label{app:complete-selfgene}
\begin{figure}[tb]
   \centering
   \includegraphics[scale=0.95]{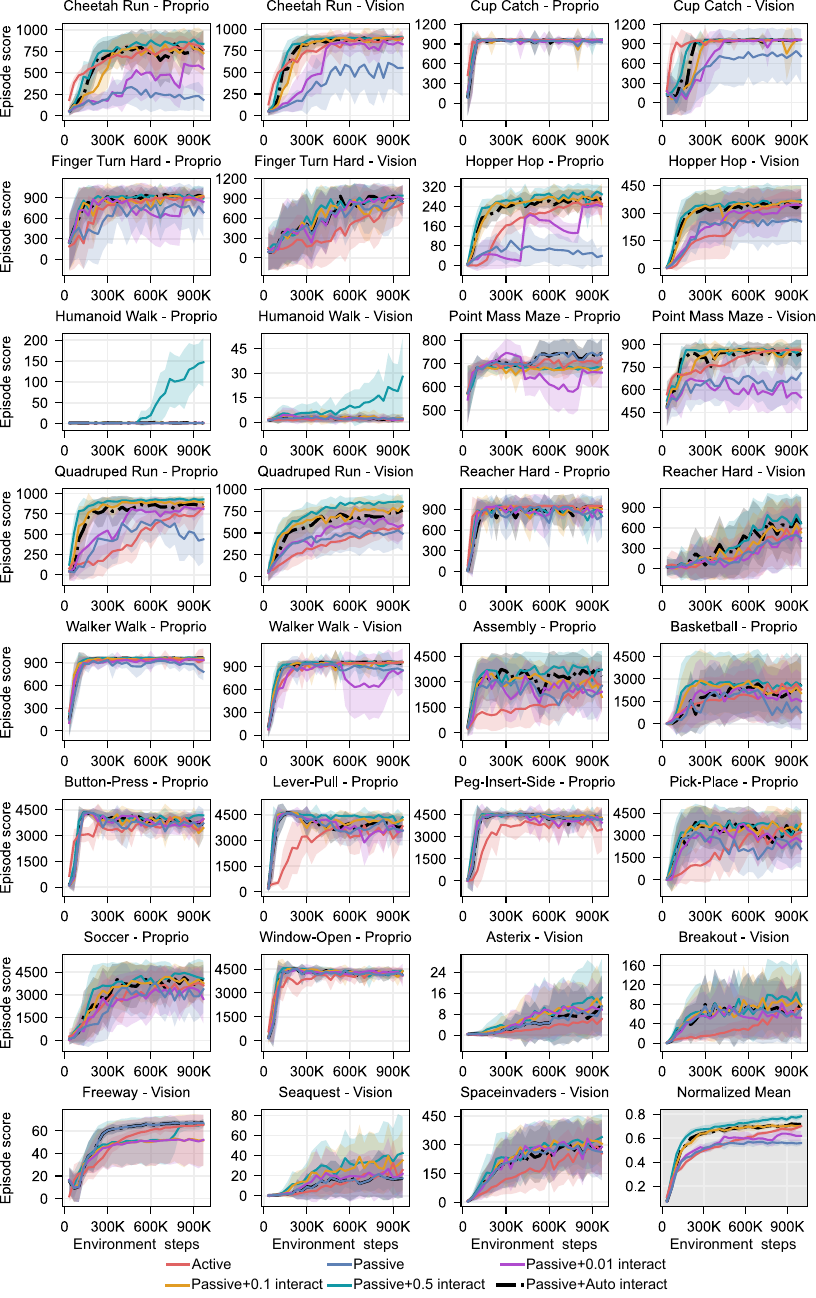}
   \caption{\textbf{Episode score of 31 tasks.} In the last subplot, we show an additional normalized mean result across tasks.}
   \label{interact_all_score}
   \vspace{-5pt}
\end{figure}

\begin{figure}[tb]
   \centering
   \includegraphics[scale=0.95]{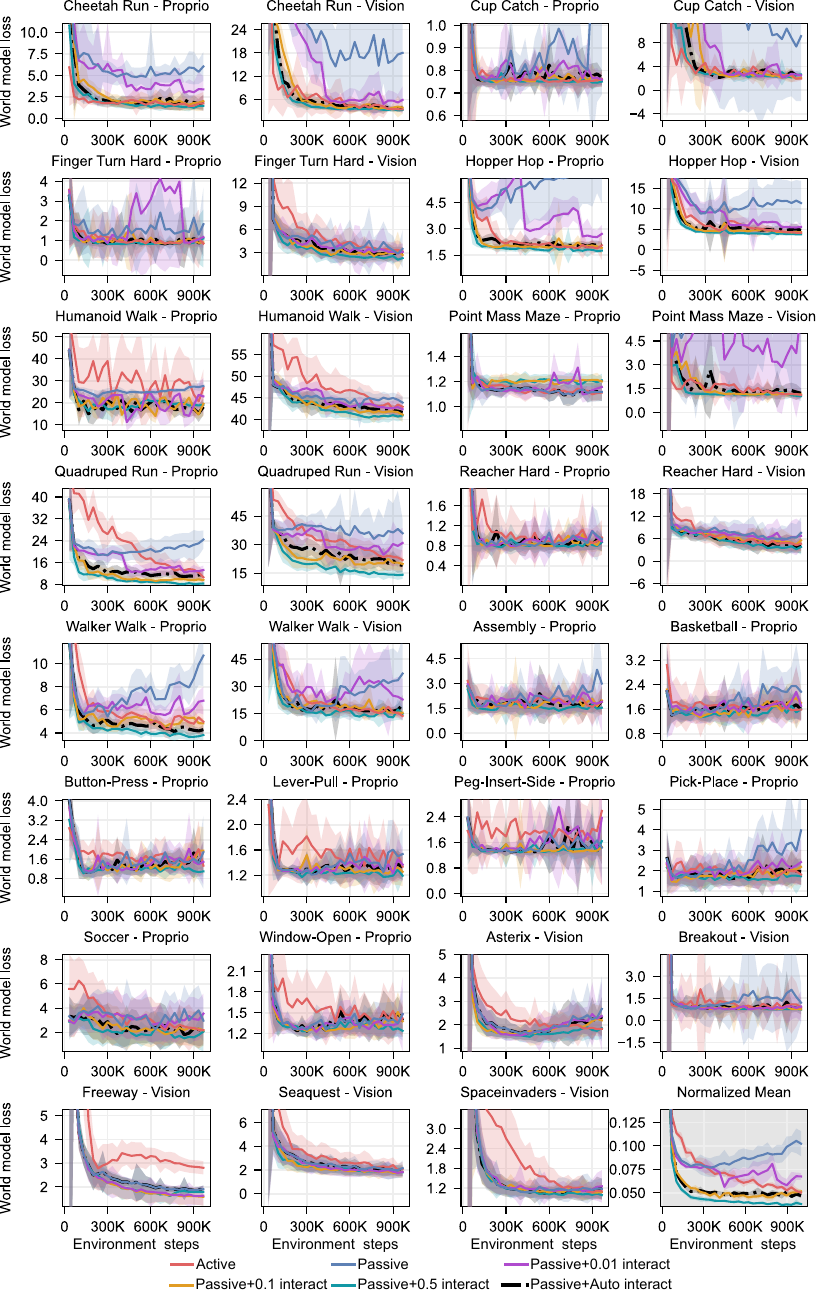}
   \caption{\textbf{World model loss of 31 tasks.} In the last subplot, we show an additional normalized mean result across tasks.}
   \label{interact_all_wm}
   \vspace{-5pt}
\end{figure}

\begin{figure}[tb]
   \centering
   \includegraphics[scale=0.95]{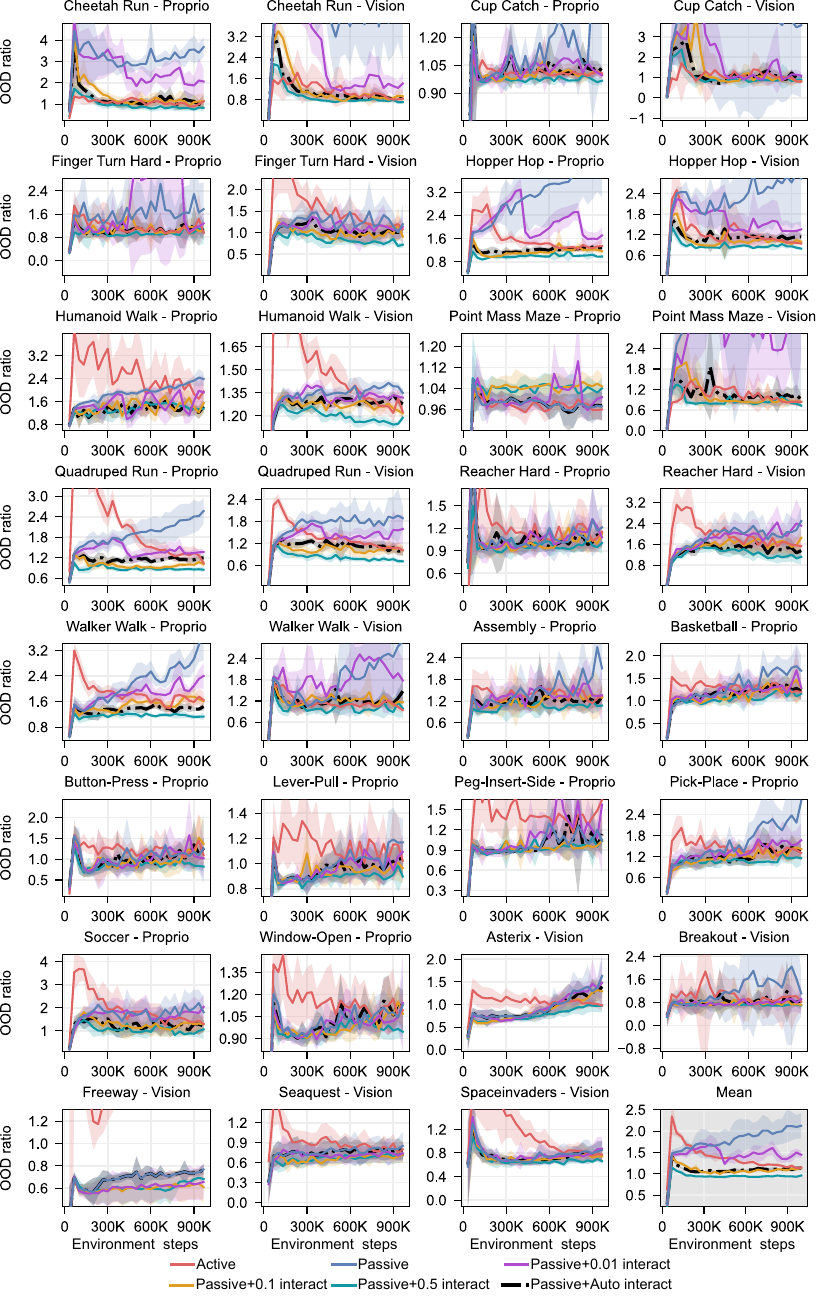}
   \caption{\textbf{\gls{ood} ratio of 31 tasks.} In the last subplot, we show an additional mean result across tasks.}
   \label{interact_all_ood}
   \vspace{-5pt}
\end{figure}

\begin{figure}[ptbp]
   \centering
   \includegraphics[scale=1.0]{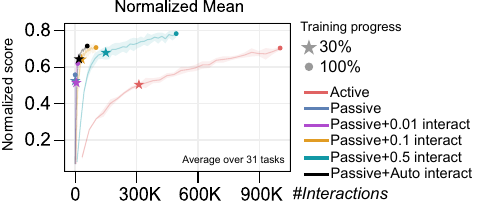}
   \caption{\textbf{Performance comparison between different \Passive agents allowed environment interaction.} The y-axis is the average normalized episode score across 31 tasks. The x-axis shows how many self-generated interaction data are added to the replay buffer. Generally, an agent with markers closest to the top left corner is the best, having the highest score and requiring minimal self-generated interaction data.}
   \label{interactions-scores}
   \vspace{-5pt}
\end{figure}

In \figref{interact_all_score}, \figref{interact_all_wm}, and \figref{interact_all_ood},  we show the complete results in 31 tasks, where we allow the \Passive agents utilize the self-generated data from environmental interaction, corresponding to the discussion in \secref{sec:self-interact}.
In \tabref{tab:added_num_AUTO}, we show how many self-generated data is added to the replay buffer by Passive+Auto interact agents.
The percentage is calculated using the number of additionally added steps divided by the total number of steps in the original replay buffer.
Together with \figref{interactions-scores}, we show that our adaptive agent \textbf{Passive+Auto interact} can converge fast and require minimal interaction data to recover the performance.

\begin{table}[pthb]
\centering
\caption{Percentage of added self-generated data by Passive+Auto interact agents}
\begin{tabular}{ll ll}
\toprule
\textbf{Task} & \textbf{Percentage (\%)} & \textbf{Task} & \textbf{Percentage (\%)} \\
\midrule
cheetah\_run-proprio & 10.44\% & walker\_walk-proprio & 18.27\% \\
cheetah\_run-vision & 6.53\% & walker\_walk-vision & 7.87\% \\
cup\_catch-proprio & 0.67\% & assembly-proprio & 8.04\% \\
cup\_catch-vision & 9.47\% & basketball-proprio & 7.16\% \\
finger\_turn\_hard-proprio & 2.53\% & button-press-proprio & 4.04\% \\
finger\_turn\_hard-vision & 3.47\% & lever-pull-proprio & 1.20\% \\
hopper\_hop-proprio & 4.31\% & peg-insert-side-proprio & 2.31\% \\
hopper\_hop-vision & 4.00\% & pick-place-proprio & 9.82\% \\
humanoid\_walk-proprio & 17.78\% & soccer-proprio & 14.93\% \\
humanoid\_walk-vision & 3.60\% & window-open-proprio & 1.47\% \\
point\_mass\_maze-proprio & 0.00\% & asterix-vision & 2.68\% \\
point\_mass\_maze-vision & 4.62\% & breakout-vision & 1.86\% \\
quadruped\_run-proprio & 2.53\% & freeway-vision & 0.00\% \\
quadruped\_run-vision & 2.93\% & seaquest-vision & 0.07\% \\
reacher\_hard-proprio & 2.27\% & spaceinvaders-vision & 0.47\% \\
reacher\_hard-vision & 20.31\% & \textbf{Average} & \textbf{5.67\%}\\
\bottomrule
\end{tabular}
\label{tab:added_num_AUTO}
\end{table}

\subsubsection{Results of Explorative Agents}\label{app:complete-expl}
\begin{figure}[tb]
   \centering
   \includegraphics[scale=0.95]{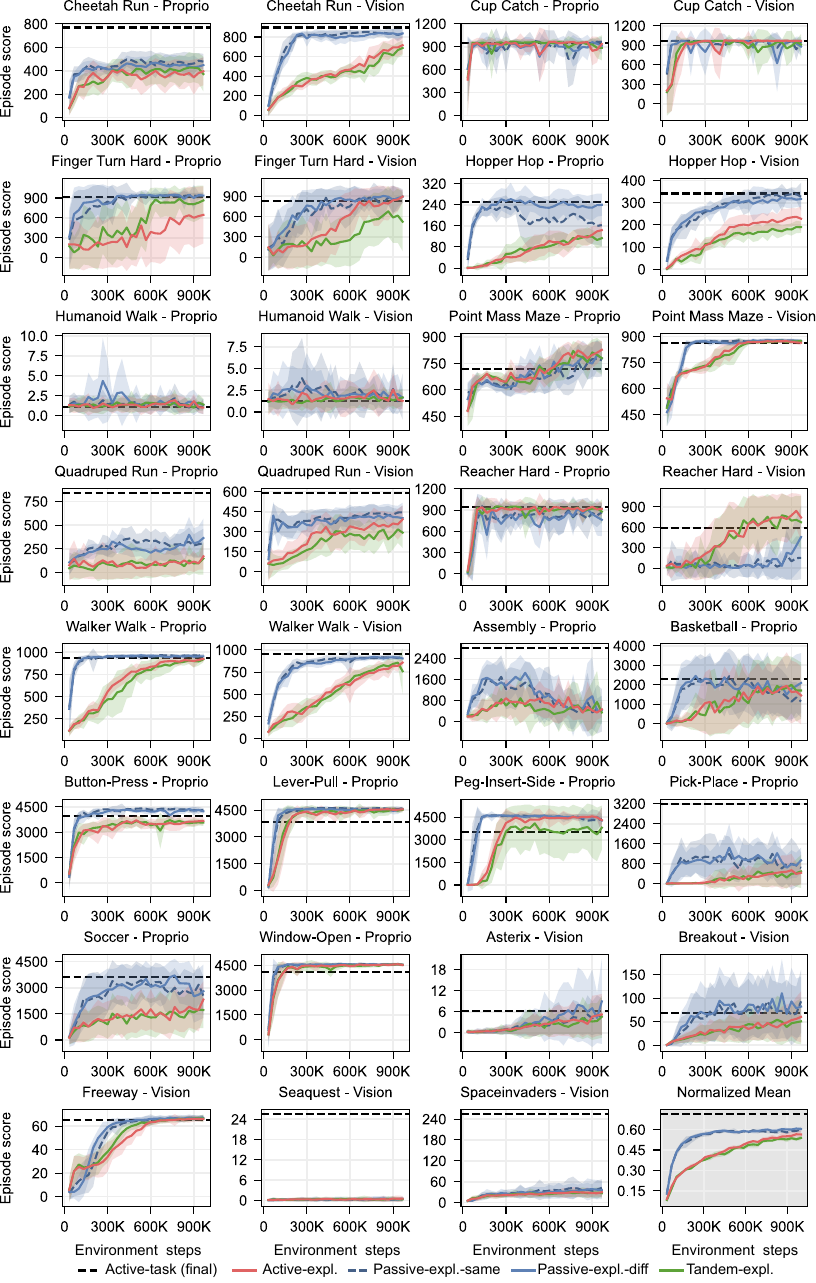}
   \caption{\textbf{Episode score of 31 tasks using agents with pure exploration rewards.} We also show the final performance of a task-oriented \Active agent as the baseline in black dashed horizontal lines. In the last subplot, we show an additional normalized mean result across tasks.}
   \label{expl_all_score}
   \vspace{-5pt}
\end{figure}

\begin{figure}[tb]
   \centering
   \includegraphics[scale=0.95]{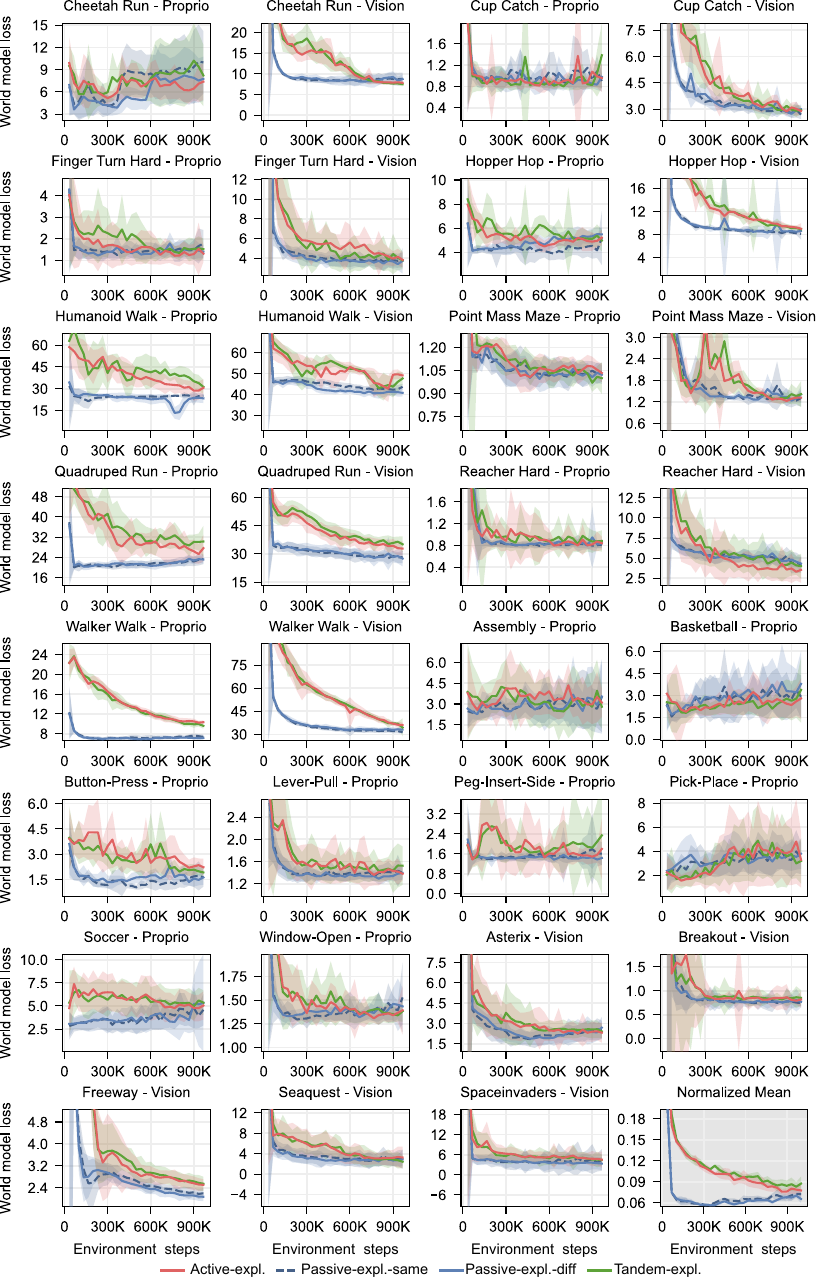}
   \caption{\textbf{World model loss of 31 tasks using agents with pure exploration rewards.} In the last subplot, we show an additional normalized mean result across tasks.}
   \label{expl_all_wm}
   \vspace{-5pt}
\end{figure}

In \figref{expl_all_score} and \figref{expl_all_wm},  we show the complete results in 31 tasks using agents with pure exploration rewards, corresponding to the discussion in \secref{sec:exploration}.
The \Passive agent initialized using the same seed for the world model and policy network as the \Active agent is marked with a suffix ``-same'', while the different model initialization is marked with ``-diff''.

\subsubsection{Results of Value Function Estimation}\label{app:complete-valuefn}
\begin{figure}[tb]
   \centering
   \includegraphics[scale=0.95]{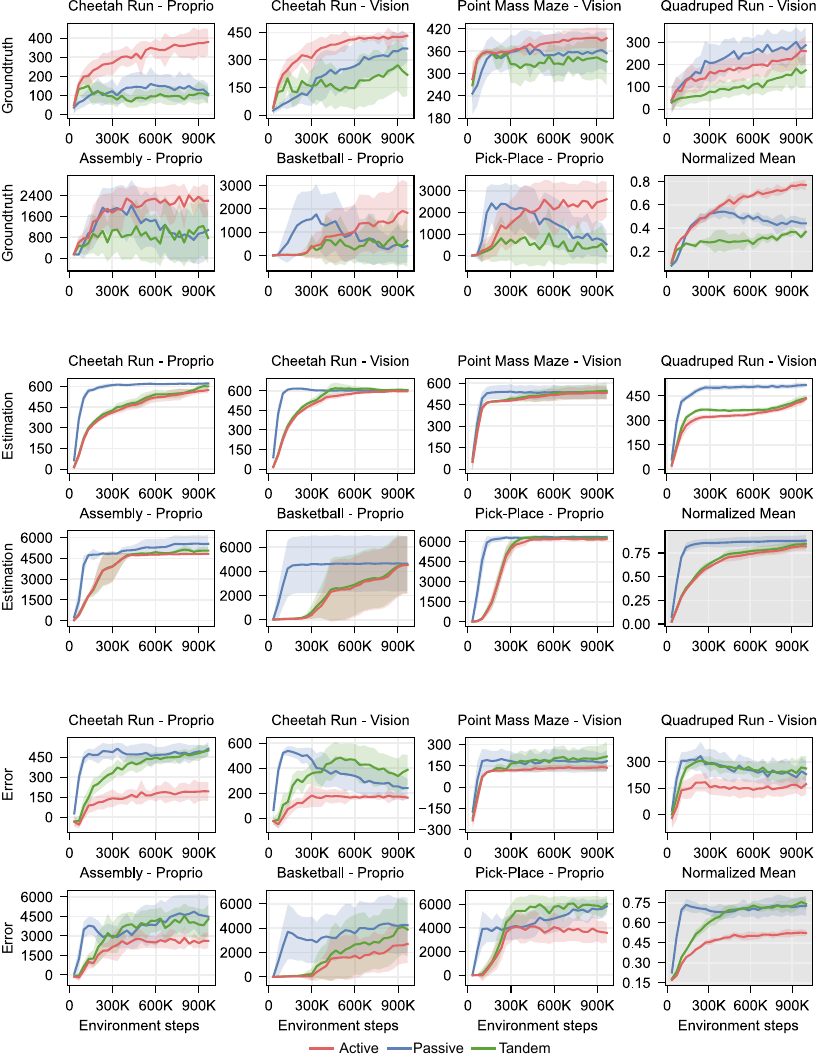}
   \caption{\textbf{Value function estimation of 7 tasks.} The value function $V(s)$ is calculated on the initial state of each agent's collected trajectory, which should reflect the actual discounted rewards accumulated across the trajectory. The ground truth value is computed using Monte Carlo estimation from one sample trajectory. The error is computed by subtracting the ground truth value from the estimated value. In the last subplot, we show an additional normalized mean result across tasks.}
   \label{discrepancy_all}
   \vspace{-5pt}
\end{figure}

In \figref{discrepancy_all}, we show the complete results in 7 tasks corresponding to the discussion in \secref{sec:discrepancy}.
The \Passive and \Tandem agents are initialized using a different seed for the world model and policy network from the \Active agent.

\subsubsection{Outliers in Complete Results}\label{app:outlier-complete-res}
Here, we briefly discuss a few task results that are unexpected or do not match the general trend and are not fully covered in the main body of the paper. A detailed analysis is beyond the scope of this work and is left for future investigation.

\paragraph{Among the results of agents trained on task-specific data in \secref{app:complete-task}}
The world model loss of the Passive\_same agent in \emph{Freeway - Vision} shows a clear mismatch with its performance degradation \wrt its \Active counterpart (\ie, low model loss despite low task performance).
This discrepancy may be attributed to the characteristics of the MinAtar domain, where an episode terminates or the agent is reset to an initial state immediately upon a failure condition, rather than continuing for a fixed length without interruption as in DMC or MetaWorld.  
In the Freeway task, the agent must cross roads while avoiding fast-moving vehicles. 
Due to the high speed and density of the cars, a failure condition (\ie, being hit) can be easily triggered.  
As a result, once the agent enters \gls{ood} states, it may be immediately reset to a familiar initial state before the world model loss can accumulate.
This reset mechanism suppresses the average world model loss, making it appear low despite poor performance.

\paragraph{Among the results of agents trained on pure exploration data in \secref{app:complete-expl}}
The \Passive agent in \emph{Reacher Hard - Vision} performs significantly worse than the \Active counterpart even when trained on exploration data.
Since this task only gives sparse rewards and the replay buffer is filled with trajectories irrelevant to the task, it is likely that the agent struggles to extract useful learning signals to perform effective policy training, especially in the early stage. 

In tasks such as \emph{Finger Turn Hard - Vision} and \emph{Peg Insert Side - Proprio}, the \Tandem agents show lower performance compared to the \Active agents. 
However, the scores are still close to the variance range of the \Active agents, suggesting that the difference may stem from seed-level randomness.
Alternatively, it could reflect a slight benefit from self-correction in the \Active agents—even under an exploration setting.
Although the task policy is not directly involved in data collection (which is guided by an exploration policy), both policies share the same world model.
Therefore, the self-correction effect driven by the exploration policy may still indirectly benefit the task policy through this shared model, with the extent depending on the task characteristics.

Besides, the world model loss of the \Tandem agent in tasks \emph{Finger Turn Hard - Vision} and of the \Passive agent in \emph{Reacher Hard - Proprio} is less indicative of their performance degradation compared to the \Active agent (\ie, the model loss remains low despite low task performance).
As mentioned in \secref{sec:exploration}, the broad state-space coverage provided by pure exploration data improves the generalization of the world model, yielding low loss even in regions far from task-relevant trajectories.
Therefore, even if performance degradation still happens due to a sparse-reward setting and insufficient task-related information in pure exploration data, its low-reward trajectory may still have low world model loss.

\paragraph{Among the results of agents adding self-generated data in \secref{app:complete-selfgene}}
In tasks such as \emph{Hopper Hop - Proprio}, \emph{Point Mass Maze - Proprio}, and \emph{Walker Walk - Vision}, the Passive agents adding 1\% interaction data (Passive+0.01 interact) exhibit large training fluctuations.
In some cases, they even underperform compared to the \Passive agents without any self-generated data.
The fluctuations may stem from the sparse addition of self-interaction data: performance rises when new interaction data is added, but declines as the model begins to overfit on the static dataset during periods without new additions.
A possible compounding factor for the worse performance is that the added self-generated trajectories—collected using a different policy—may lie outside the original training distribution. 
Such trajectories may inject noise into the world model training by providing inconsistent training signals.
Due to their small proportion (1\%), they fail to offer meaningful self-correction and instead destabilize the learning process to varying degrees, depending on the task.

\paragraph{Among the results of value function estimation in \secref{sec:discrepancy}}
In an earlier standalone offline test using a final checkpoint of the \emph{Pick-Place - Proprio} task, we observed one mismatch between the value function error and the task performance of the \Passive agent (\ie, low value prediction error despite a low task score compared to the \Active counterpart).
However, after rerunning the task with three random seeds and collecting statistics over time, the results aligned with our main claims.
This suggests that the earlier observation was likely due to randomness or variance.

\paragraph{Among the results of suboptimal and mixed dataset in \secref{sec:ood-app-detail}}
While mixing expert trajectories is generally beneficial for offline training, we observe an exception in \emph{Point Mass Maze - Vision} in \figref{split}, where the Passive\_mixed agent underperforms the Passive\_suboptimal agent in the second half of training.
A possible reason is that the \Active agent learns to solve the task early (already by the end of the first half of the replay buffer), so the second half mostly contains repetitive, task-specific trajectories with little new information—as reflected in its consistently high performance and low variance.
As a result, mixing samples from both halves may actually reduce the overall diversity compared to using only the first half, leading to narrower state-space coverage and increasing \gls{ood} risk.
This is also consistent with the slightly higher world model loss observed in Passive\_mixed compared to Passive\_suboptimal.

\clearpage % Forces figures to appear before 
\section{Additional Discussions}
\subsection{Connection to the Original Tandem RL Paper~\citep{ostrovski2021difficulty}}

\paragraph{Motivation: addressing performance degradation in model-based offline training} 
Our work is motivated by the performance degradation often observed when training model-based agents offline. 
While we draw inspiration from the experimental setup of the original Tandem RL paper~\citep{ostrovski2021difficulty}, their focus is on model-free methods (specifically DQN), and their analysis centers on the policy/value function. 
In contrast, we focus on a deeper and more intricate level of analysis—specifically, the interplay between the world model and the policy, which lies at the core of model-based RL.
Our study complements the findings in~\citet{ostrovski2021difficulty} by extending the tandem framework to the model-based domain, where the sources of degradation are inherently different and less well-understood.
Although such degradation and its solutions are sometimes assumed intuitively in the model-based RL community, it has lacked empirical backing—our work fills this gap with systematic and extensive experiments.

\paragraph{Experimental design: similar principles, adapted to a different focus}
Our experimental design is inspired by the controlled setup of Tandem agents proposed in~\citet{ostrovski2021difficulty}, where training data stream is held constant to isolate the effect of other factors.
We follow the same underlying principle of controlled experimentation to identify causal factors in performance degradation, \eg, by adopting a similar Tandem agent setup.
However, we adapt this methodology to the model-based setting, designing new test cases -- \eg fix the same world model to examine whether degradation arises from world model learning alone (\secref{subsubsec:why-ood}).
Moreover, we conduct experiments across a broader range of environments beyond the original Atari focus, evaluating the generality of our findings in diverse model-based RL scenarios.

\subsection{Contextualization in Similar Work}

Most closely related studies on performance degradation in offline RL focus on model-free methods~\citep{yarats2022exorl, ostrovski2021difficulty, mediratta2024gengap}. Therefore, their conclusions should not be assumed to directly transfer to model-based settings, where the interaction between the world model and policy can introduce different failure dynamics.

Even in model-based works~\citep{Agnostic}, the analysis remains limited—\eg it could benefit from a deeper investigation into the respective roles of the world model and the policy, as well as from broader evaluation beyond a single-domain setting.
Other solution-oriented works like~\citet{MOReL} mainly propose algorithmic solutions rather than understanding degradation from a data-centric perspective.

Our goal is not algorithmic novelty, but to fill this analytical gap through controlled and extensive experiments, offering a clearer understanding of degradation in model-based offline RL and identifying what kinds of data can help mitigate it.

\end{document}